\documentclass[10pt,twocolumn,letterpaper]{article}

\usepackage{cvpr}              

\definecolor{cvprblue}{rgb}{0.21,0.49,0.74}
\usepackage[pagebackref,breaklinks,colorlinks,allcolors=cvprblue]{hyperref}
\usepackage{bbm}
\usepackage{multirow}
\newcommand{\cmark}{\ding{51}}
\newcommand{\xmark}{\ding{55}}
\usepackage{pifont}
\usepackage{algorithm}
\usepackage{amsmath}
\usepackage{amssymb}
\usepackage{pifont}
\usepackage{booktabs}
\usepackage{makecell}
\usepackage{colortbl}
\usepackage{algpseudocode}
\PassOptionsToPackage{table}{xcolor}
\newcommand{\code}[1]{\textcolor[RGB]{88,142,94}{#1}}
\usepackage{graphicx}
\usepackage{adjustbox}

\usepackage{etoolbox}

\makeatletter
\newcommand{\secondcorrespondingauthor}{%
  \@fnsymbol{2}
}
\makeatother

\def\confName{CVPR}
\def\confYear{2025}

\title{Multi-Granularity Class Prototype Topology Distillation for Class-Incremental Source-Free Unsupervised Domain Adaptation}

\author{Peihua Deng$^{1,5}$\thanks{This work is done during the intern in VIPL group, ICT, CAS.}, 
Jiehua Zhang$^2$, 
Xichun Sheng$^3$, 
Chenggang Yan$^{1}$\thanks{Corresponding authors.}, 
Yaoqi Sun$^1$, 
Ying Fu$^4$, 
Liang Li$^{5}$\textsuperscript{\secondcorrespondingauthor}\\
$^1$Hangzhou Dianzi University, $^2$Xi'an Jiaotong University, $^3$Macao Polytechnic University\\
$^4$Beijing Institute of Technology, $^5$Institute of Computing Technology, Chinese Academy of Sciences\\
\small
{\texttt{peihuadeng@hdu.edu.cn}, 
\texttt{jiehua.zhang@stu.xjtu.edu.cn}, 
\texttt{cgyan@hdu.edu.cn}, 
\texttt{liang.li@ict.ac.cn}}
}

\begin{document}
\maketitle
\begin{abstract}
This paper explores the Class-Incremental Source-Free Unsupervised Domain Adaptation~(CI-SFUDA) problem, where the unlabeled target data come incrementally without access to labeled source instances. This problem poses two challenges, the interference of similar source-class knowledge in target-class representation learning and the shocks of new target knowledge to old ones. 
To address them, we propose the Multi-Granularity Class Prototype Topology Distillation~(GROTO) algorithm, which effectively transfers the source knowledge to the class-incremental target domain. 
Concretely, we design the multi-granularity class prototype self-organization module and the prototype topology distillation module. First, we mine the positive classes by modeling accumulation distributions. Next, we introduce multi-granularity class prototypes to generate reliable pseudo-labels, and exploit them to promote the positive-class target feature self-organization. Second, the positive-class prototypes are leveraged to construct the topological structures of source and target feature spaces. Then, we perform the topology distillation to continually mitigate the shocks of new target knowledge to old ones. 
Extensive experiments demonstrate that our proposed method achieves state-of-the-art performance on three public datasets. Code is available at https://github.com/dengpeihua/GROTO.
\end{abstract}    
\section{Introduction}
With the development of deep learning~\cite{Liu2022EntityEnhancedAR,zhang2024deep,zhang2024speaker,10433795}, transfer learning has made great progress. Unsupervised domain adaptation~(UDA)~\cite{10124696,10669385,cui2024stochastic,zhou2024unsupervised} relies on labeled source data to mitigate the domain gap. However, in some practical applications, it is infeasible to access source data when training on the target domain, while only the pre-trained source model and unlabeled target data are provided, known as source-free unsupervised domain adaptation~(SFUDA)~\cite{qu2022bmd,karim2023c,mitsuzumi2024understanding,xu2025unraveling}.

\begin{figure}[t]
  \centering
\includegraphics[width=1.0\linewidth]{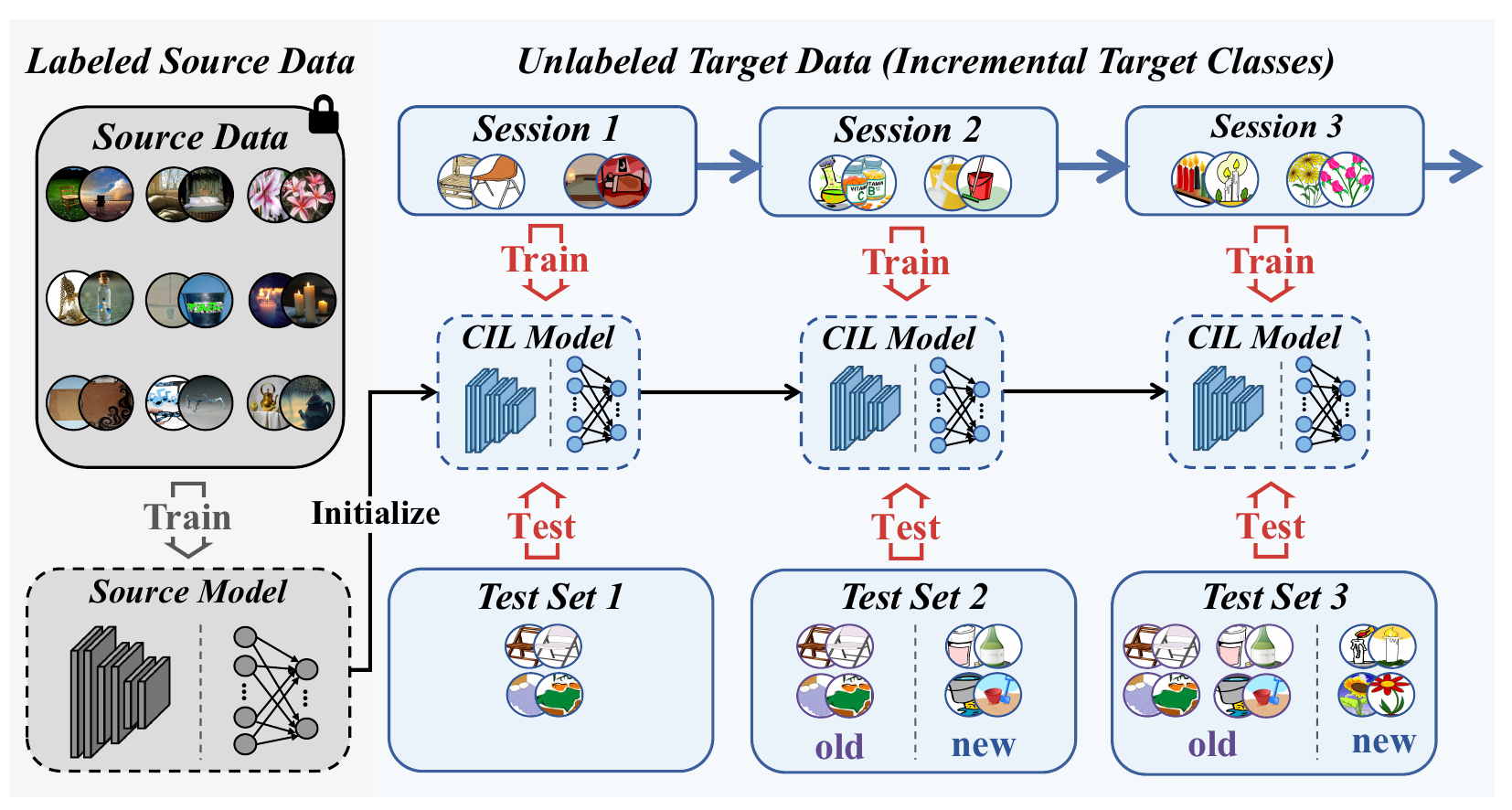}
  \caption{An illustration of Class-Incremental Source-Free Unsupervised Domain Adaptation~(CI-SFUDA) problem, where the labeled source data contain all classes while unlabeled target data come incrementally without access to source instances, and the previously learned target data are unavailable for later adaptations.}
  \label{Figure 1}
  \vspace{-8pt}
\end{figure}

To facilitate knowledge transfer, most SFUDA works adapt target distributions with self-training methods, where pseudo-labels are generated by a non-parameter classifier based on the similarity between image features and class prototypes.
However, these works are established on the identical label space between source and target domains. In fact, the model often trains on a labeled large-scale dataset and then transfers knowledge to the unlabeled small-scale target domain~\cite{cao2019learning,li2024principal}. 
In such a non-identical label space scenario, traditional SFUDA works may push the unlabeled target data toward similar source-class clusters outside the target label space. 
Although the latest source-free universal domain adaptation~(SF-UniDA) works can adapt target domain distribution in the case of inconsistent label space through feature decomposition and knowledge distillation methods~\cite{qu2024lead,tang2024sourcefreedomainadaptationfrozen}, they assume the availability of all target data in advance. 
In a dynamically changing environment, all the target data cannot be easily collected at once and always come in streams of different classes~\cite{kundu2020class,niu2022efficient,lin2022prototype}. 
In such a case, existing SF-UniDA works cannot continually adapt to class shifts. When new class data arrive, they require training from scratch, consuming substantial time and storage resources~\cite{kundu2020class,lin2022prototype}.
For this issue, the unsupervised class incremental learning~(UCIL) work~\cite{liu2024exploiting} can learn new class knowledge with a strong pretrained model while retaining the old class ones. However, it struggles to address domain distribution shifts. 
Therefore, the pretrained source model is expected to continually adapt the unlabeled target data with domain and class shifts without re-training.

In this paper, we explore the \textit{Class-Incremental Source-Free Unsupervised Domain Adaptation}~(CI-SFUDA) problem, where the target data come incrementally as sequential sessions without available source instances, while the target classes of each session derive from a subset of the source label space as shown
in~\Cref{Figure 1}.
Therefore, CI-SFUDA faces two challenges: (1) the interference of similar source-class knowledge in target-class representation learning. The target label space of each session occupies only a subspace of the source one, unlabeled target data may be incorrectly assigned pseudo-labels from similar source classes outside the session, resulting in biased target representations. 
(2) The shocks of new target knowledge to old ones. 
Previously learned data are unavailable for new target class adaptation, thus when adapting to new classes, the optimizer updates model weights to adjust towards new features and patterns, leading to the forgetting of old knowledge.

To address them, we propose the Multi-\textbf{G}ranularity Class P\textbf{RO}totype \textbf{TO}pology Distillation~(GROTO) algorithm, which aims to continually adapt a source model to incremental target classes across different sessions.
Firstly, we design the multi-granularity class prototype self-organization module to mitigate the interference of the inconsistent label space in target representation learning at each session. 
Specifically, we introduce the hybrid knowledge-driven positive class mining method, which models the accumulation distributions of source similarity and target probability to align label spaces. 
Then, we compute multi-granularity class prototypes, which leverage correctly classified unlabeled target data well to reduce noise from hard-transfer data, thus generating reliable pseudo-labels.
With pseudo-labeled positive-class data, we perform the target feature self-organization to pull the same-class ones closer and push the different-class ones farther apart.

Secondly, we propose the prototype topology distillation module to continually reduce the shocks of new target knowledge to old ones. In detail, we compute positive-class prototypes to construct the topological structures of source and target feature spaces. Then, we design the losses to perform point-to-point distillation, which maintains the consistency between the source knowledge learned from all classes at once and the target knowledge gradually accumulated across multiple sessions, thus reducing the decision boundary overfitting to new target classes.

Experiments on three benchmark datasets show that our algorithm outperforms state-of-the-art methods. In summary, the main contributions of this paper are as follows:

\begin{itemize}
\item We explore the class-incremental source-free unsupervised domain adaptation~(CI-SFUDA) problem, and propose the multi-granularity class prototype topology distillation~(GROTO) algorithm.
\item We design the multi-granularity class prototype self-organization module to mitigate the interference from similar source-class knowledge to the target-class representation learning.
\item We design the prototype topology distillation module to reduce the shocks of new target knowledge to old ones.
\end{itemize}

\label{sec:Introduction}

\section{Related Work}
\textbf{Source-free Domain Adaptation~(SFDA).} 
With the advancement of artificial intelligence in recent years~\cite{ye2022unsupervised,cvpr/gait3d_v1,tian2023transformer,10607969}, domain adaptation has also made great progress. 
SFDA works aim to transfer a pretrained source model to the unlabeled target domain~\cite{qu2022bmd,karim2023c,qu2024lead,zhou2024source,liu2024source}. BMD~\cite{qu2022bmd} exploits the multicentric dynamic prototype method to generate robust pseudo-labels, but its numbers of empirical prototypes are fixed and always different in the various datasets. C-SFDA~\cite{karim2023c} introduces curriculum learning to filter noisy pseudo-labels. However, they struggle to handle adaptation problems involving label space discrepancies.
Recent SF-UniDA works~\cite{qu2023upcycling,tang2024unified,qu2024lead} largely address the adaptation of inconsistent label space only with the source model and unlabeled target data. 
GLC~\cite{qu2023upcycling} employs clustering to distinguish domain-public and domain-private data. 
The statistical distribution-based and causality-driven feature decomposition works~\cite{qu2024lead,tang2024unified} reduce the interference of irrelevant feature information.
In practical scenarios where the target data come in a streaming manner with different classes, these works fail to address it.

\textbf{Class-Incremental Learning~(CIL).} CIL~\cite{tao2020topology,liu2022model,E220396,he2024gradient,meng2024diffclass} faces a scenario where different class data arrive incrementally as sequential sessions, and previously learned data are no longer available at new sessions. 
Memory-based works store representative exemplars of previous tasks and then perform memory replay~\cite{kang2022class,lin2023class} to mitigate the catastrophic forgetting~\cite{wu2019large}. Non-memory works~\cite{huang2024ovor,huang2024class} deal with new tasks by altering network structures or optimizing parameters.
Most CIL works are supervised, but it is difficult to obtain complete ground-truth labels as new data arrive. Therefore, the UCIL work~\cite{liu2024exploiting} leverages powerful pre-trained models to capture comprehensive feature representations from unlabeled data and models the distributions of old and new classes. Then, it performs granularity alignment while reducing overlap between the old and new classes, thereby mitigating the forgetting problem. However, it cannot adapt to shifts in data distribution within the same class. Our GROTO can simultaneously deal with the domain and class shifts in the CI-SFUDA. 

\label{sec:RelatedWork}

\begin{figure*}[tp]
  \centering
  \includegraphics[width=1.0\linewidth]{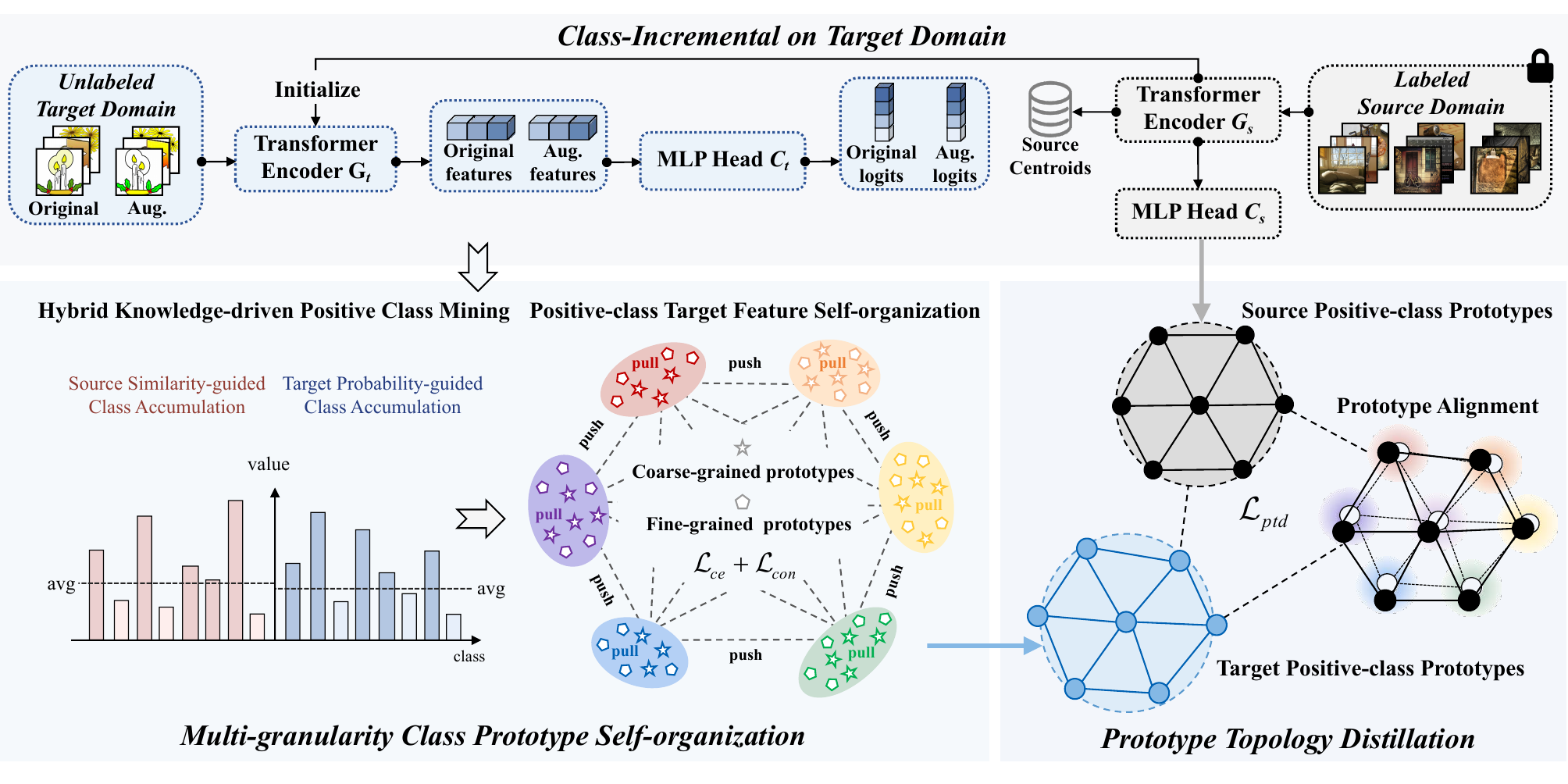}
  \caption{An overview of the GROTO algorithm, it includes two modules: 1) Multi-granularity class prototype self-organization: we mine the positive classes by modeling the source similarity and target probability accumulation distributions, and then promote the positive-class target features self-organization based on the multi-granularity class prototypes via $\mathcal{L}_{ce}$ and $\mathcal{L}_{con}$. 
2) Prototype topology distillation: we compute the positive-class prototypes to distillate the topological structures of source and target feature spaces via $\mathcal{L}_{ptd}$.}
  \label{Figure 2}
  \vspace{-15pt}
\end{figure*}
\section{Proposed Method}
\subsection{Problem Formulation}
Let $\mathcal{D}_s=\{(x_j^s,y_j^s)|y_j^s\in Y^s\}_{j=1}^{n_s}$ denotes the source domain with ${n}_{s}$ source data, where $Y^s$ is the source label set and the number of source classes is $|Y^s|=K$. 
We need to adapt the source model $M_s$ to the sequential training datasets $D^1, D^2,\cdots, D^t, D^{t+1}$, and build a unified target model $M_t$, where $D^t=\{(x_i)\}_{i=1}^{n_t}$ is the unlabeled target domain of the $t$-th session with $n_{t}$ samples, $G_t$ and $C_t$ are the feature extractor and classifier of $M_t$, $G_s$ and $C_s$ are the feature extractor and classifier of $M_s$. $Y^{p}$ and $Y^{q}$ are disjoint target label sets at the $p$-th and $q$-th target sessions, \textit{s.t.} $\forall p\neq q,Y^p\cap Y^q=\text{Ø}$. Moreover, $Y^{t}$ at each session $t$ is a subset of the source label set, \textit{i.e.} $Y^t\subset Y^s$. At session $t+1$, the target model $M_t$ is learned from $D^{t+1}$ without the old-class training sets $D^1,D^2,\cdots,D^t$. Then, $M_t$ is evaluated on the concatenated set of all the seen sets $\bigcup_{i=1}^{t+1}D^i$.

\subsection{Overall Framework}
The overall framework of GROTO is shown in \Cref{Figure 2}. In CI-SFUDA, the target classes of different sessions are non-overlapping subsets of the source label space. Domain-shared classes are considered positive classes, and the remaining classes in the label spaces are considered negative classes. \textbf{Firstly}, to reduce the interference of similar source-class knowledge in the target-class representation learning, we design the multi-granularity class prototype self-organization module, identifying domain-shared positive classes and then conducting target feature self-organization. In detail, we mine positive classes of new sessions by modeling the source similarity and target probability accumulation distributions. Then, we generate pseudo-labels for target data by identifying the coarse-grained and fine-grained class prototypes. With these reliable pseudo-labels, we conduct the target feature self-organization with the cross-entropy loss $\mathcal{L}_{ce}$ and contrastive loss $\mathcal{L}_{con}$. 
\textbf{Secondly}, to continually mitigate the shocks of new target knowledge to old ones, we propose the prototype topology distillation module. We compute positive class prototypes to construct the topological structures of feature spaces, then perform point-to-point distillation between the topological structures of the source and the target via the prototype topology distillation loss $\mathcal{L}_{ptd}$.

\subsection{Multi-granularity Class Prototype Self-organization}
The mainstream paradigm is the self-training framework, where pseudo-labels are generated for unlabeled target data to facilitate model adaptation. However, the target classes come incrementally in CI-SFUDA, and the label set $Y^t$ of each session $t$ is a subset of the source one $Y^s$. Incorrect pseudo-labels out of $Y^t$ may be assigned to unlabeled target data at the $t$-th session, pushing target features to similar source-class clusters outside the target label space $Y^t$.
To solve this problem, we propose the multi-granularity class prototype self-organization module that consists of two steps: 1) hybrid knowledge-driven positive class mining~(HKPCM), 2) positive-class target feature self-organization~(PTFS).

\noindent\textbf{Hybrid Knowledge-driven Positive Class Mining.} 
To identify positive classes, the hybrid knowledge-driven positive class mining method with the accumulation distributions of source similarity and target probability is proposed. Specifically, for the source similarity-guided class accumulation distributions, we first load the stored source feature centroids $s=[s_1,...,s_K]$ of all the $K$ source classes. Next, we use the source feature extractor $G_s$ to generate features $g=[g_{1},...,g_{n_{t}}]$ for $n_{t}$ target data at the $t$-th session, and calculate the similarity matrix ${R} \in \mathbb{R}^{n_t \times K}$ by:
\begin{equation}
    R={g^T}\cdot s,
\end{equation}
where each element $r_{ij} \in {R}$ represents the similarity between the $i$-th target data and the ${j}$-th source centroid. We normalize $r_{ij}$ with the softmax normalization and compute the average similarity $S_k$ for each source class $k$. Then, we identify the positive classes by $S_{k} > \frac{1}{K}\sum_{j=1}^{K}S_j$, as shown in the red histogram in \Cref{Figure 2}.

For the target probability-guided class accumulation distributions, we generate the prediction for each target data $x_i$ of the $t$-th session to obtain the cumulative probability $P_k$ for each source class $k$ in $Y^s$:
\begin{equation}
P_k=\sum_{i=1}^{n_t}C_s(G_s(x_i)).
\end{equation}
We normalize the cumulative predicted probability $P_k$ of each source class $k$ with the min-max normalization and identify the positive classes by $P_{k}>\frac{1}{K}\sum_{j=1}^KP_j$, as shown in the blue histogram in \Cref{Figure 2}.

With the pseudo label assigned in subsequent stages, unique pseudo-labels evolve as the positive classes.

\noindent\textbf{Positive-class Target Feature Self-organization.} To generate reliable pseudo-labels, we first identify the coarse-grained and fine-grained prototypes. Then we conduct target feature self-organization via the cross-entropy loss and contrastive loss to make the intra-class features more compact and inter-class features more separable. 

\textit{Coarse-grained prototypes} $O_c$ include the source and target coarse-grained ones of positive classes.
The source classifier weights $\mu=[\mu_1,...,\mu_{N}]$ of $N$ positive classes are considered as the source coarse-grained prototypes, which denote the stable convergence points of each positive class on the source domain. Further, considering the domain gap, we identify target coarse-grained prototypes to provide the representation of target-class central tendencies, which are representative data with high similarity to target feature centroids of positive classes.
In detail, for each positive class $n$, the class centroid $c_n$ is calculated by averaging features with the same initial pseudo-label ${\arg\max}_n{C_t(G_t(x_i))}=n$ by: 
\begin{equation}
c_n=\frac1{\mid S_n\mid}\sum_{x_n\in S_n}G_t(x_n),
\end{equation}
where $S_n$ is the set of all the target data that pseudo-label is $n$, $x_n$ is the data in $S_n$. For each data $x_n$, we calculate its cosine distance with $c_n$ by $d(x_n,c_n)=1-\frac{G_t(x_n)\cdot c_n}{\|G_t(x_n)\|_2\|c_n\|_2}$, and compute the average cosine distance of all the data with initial pseudo label $n$ as the threshold $\tau_s$:
\begin{equation}
\tau_s=\frac1{\mid S_n\mid}\sum_{x_n\in S_n}d(x_n,c_n).
\end{equation}
We identify the features as target coarse-grained prototypes of positive class $n$ by $d(x_n,c_n) < \tau_s$.

\textit{Fine-grained prototypes} $O_f$ are the target features with reliable initial pseudo-labels, which provide detailed representations about each target class. To select the fine-grained prototypes, we first conduct general data augmentation to simulate the distribution offsets. Then we calculate the uncertainty $u_{i}$ of the prediction confidences~(\textit{conf}) for
each target data $x_i$ and its augmented data $x_i^{\prime}$ of the $t$-th session by taking the standard deviation~(\textit{std}): 
\begin{equation}
u_{i}=std(conf(x_i),conf({x_i^{\prime}})).
\end{equation}
The thresholds $\tau_c$ and $\tau_u$ are computed as: 
\begin{equation}
\tau_c=\frac1{2n_t}\sum_{i=1}^{n_t}\left(conf(x_i)+conf({x_i^{\prime}})\right),
\quad\tau_u=\frac1{n_t}\sum_{i=1}^{n_t}u_i.
\end{equation}
The feature of $x_i$ is considered one of the fine-grained prototypes $O_f$ by: 
\begin{equation}
conf_{avg}\left(x_i,{x_i^{\prime}}\right)>\tau_c\quad\mathrm{and}\quad u_{i}<\tau_u,
\end{equation}
where $conf_{avg}$ is the average prediction confidence of the data pair.

Let $O=O_c\cup O_f$ be the union of all the coarse-grained and fine-grained prototypes, and $o_n\in O$ be the prototypes of each positive class $n$. We compute the average of the cosine distances from each of the remaining target data $x_j$ to all the $n$-th positive-class prototypes $o_n$ as the distance of $x_j$ to the $n$-th positive-class $D(x_j,n)$. Then, we assign the pseudo-label $\bar{y}_j$ to $x_j$:
\begin{equation}
\bar{y}_j=\arg\min_n{D(x_j,n)}.
\end{equation}

Finally, we conduct target feature self-organization with the cross-entropy loss and unsupervised contrastive learning~\cite{khosla2020supervised}: 
\begin{equation}
\mathcal{L}_{ce}=-\frac{1}{B}\sum_{i=1}^{B}\bar{y}_{i}\cdot\log C_t(G_t(x_i)),
\end{equation}
\begin{equation}
\begin{gathered}
\mathcal{L}_{con}=\frac{1}{2B}\sum_{b=1}^{2B}[\ell_{2b-1,2b}+\ell_{2b,2b-1}],
\end{gathered}
\end{equation}
\begin{equation}
\begin{gathered}
\ell_{i,j}=-\log\frac{\exp(\operatorname\phi(G_t(x_i),G_t(x_i^{\prime}))/\kappa)}{\sum_{b=1}^{2B}\mathbbm{1}_{b\neq i}\exp(\operatorname\phi(G_t(x_i),G_t(x_b))/\kappa)},
\end{gathered}
\end{equation}
\begin{equation}
\mathcal{L}_{ptfs}=\mathcal{L}_{ce}+\mathcal{L}_{con},
\end{equation}
where $B$ is the batch size, $\bar{y}_{i}$ is the pseudo label of $x_i$, $\kappa$ is the temperature constant, $\phi(\cdot,\cdot)$ is the cosine similarity and $\mathbbm{1}_{b\neq i}$ is an indicator function that gives a $1$ if $b\neq i$,. By combining $\mathcal{L}_{con}$ and $\mathcal{L}_{ce}$~(denoted as $\mathcal{L}_{ptfs}$), we can make the embedding space more compact while preserving the fine-grained information about the decision boundary of each target class.

\subsection{Prototype Topology Distillation}
At a new incremental session, the optimizer updates model weights to adjust towards the features and patterns of new classes, leading to the forgetting of old knowledge.
ProCA~\cite{lin2022prototype} mitigates this problem with the memory replay method. It constructs a memory bank $\mathcal{M}=\{(m_i,\hat{y}_i,\bar{y}_i)\}_{i=1}^{N_r}$ to store a small number of representative exemplars for replaying all $N^{\prime}$ existing positive classes, where $m_i$, $\hat{y}_i$, $\bar{y}_i$ and $N_r$ respectively denote the exemplar, soft prediction, pseudo-label and the number of all exemplars. For each existing class $n^{\prime}$ in the memory bank, $n_r$ exemplars are stored, where $n_r={N_r/N^{\prime}}$. 
To maintain the relative position relationship of old and new target classes in the feature space, our GROTO also maintains a similar memory bank. To make it compatible with our HKPCM method, if the same positive class is recognized in multiple sessions, we only store the positive class prototype in the session with higher confidence in $\mathcal{M}$.
Then, We perform the exemplar replay~\cite{wu2019large} by: 
\begin{equation}
\mathcal{L}_{rep} = -\frac{1}{N_r}\sum_{i=1}^{N_r} {\hat{y}_i}^\top \log C_t(G_t(m_i)),
\end{equation}

In CI-SFUDA, classes of different sessions are not overlapping, resulting in data imbalance between the old and new classes~($N_r\ll n_t$). Thus, relying solely on the exemplar memory replay method leads the decision boundary to overfit the new dominant classes~\cite{de2021continual,li2023effective}. In our setting, the source model is trained on labeled data of all the classes at once, acquiring discriminative knowledge of all the classes. This indicates that we can utilize the rich discriminative information learned from all source classes to alleviate the decision boundary overfitting in the incremental process by aligning classifier weights between domains, thereby creating a link between transfer learning and continual learning.

In light of this, we propose the prototype topology distillation~(PTD) module, which continually reduces the shocks of new target knowledge to old ones. 
During the target incremental training process, we utilize the source topological structure to prospectively support the decision boundary construction of new and old classes at each session.
In detail, we first adopt classifier weights of the source and target model as source prototypes $\mu=[\mu_1,...,\mu_{N}]$ and the target ones $f=[f_1,...,f_{N}]$ to construct the topological structures of feature spaces for $N$ positive classes. 
As the target prototypes tend to be moved towards the neighboring source prototypes with the same label~\cite{tanwisuth2021prototype}, we design the compactness loss $\mathcal{L}_{com}$:
\begin{equation}
\mathcal{L}_{com}=\frac{1}{N}\sum_{j=1}^{N}\sum_{i=1}^{N}d(\mu_{i},f^T_{j})\frac{p(f_{i})\mathrm{exp}(\mu_{i}f_{j}^{T})}{\sum_{i^{\prime}=1}^{N}p(f_{i^{\prime}})\mathrm{exp}(\mu_{i^{\prime}}f_{j}^{T})},
\end{equation}
where $p(\cdot)$ is the target proportion. The unnormalized likelihood term $\mathrm{exp}(\mu_{i}f_{j}^{T})$ measures the similarity of the $i$-th source and $j$-th target prototypes. The cosine distance $d(\cdot,\cdot)$ measures the point-to-point distillation cost of each target prototype to each source one, allowing the model to make more refined adjustments to the target prototypes. 
Minimizing $\mathcal{L}_{com}$ means the distillation cost minimization, encouraging target prototypes to be closer to the source ones they are supposed to be mapped to, thereby enhancing the intra-class compactness of the same class between domains.

But minimizing $\mathcal{L}_{com}$ alone for target prototypes may lead to the same tendency of discarding some patterns~\cite{zheng2021comparing}. Thus, we further design the separability loss $\mathcal{L}_{sep}$:
\begin{equation}
\mathcal{L}_{sep}=\sum_{i=1}^Np(f_i)\sum_{j=1}^Nd(\mu_i,f^T_j)\frac{\exp(\mu_if_j^T)}{\sum_{j^{\prime}=1}^N\exp(\mu_if_{j^{\prime}}^T)},
\end{equation}
where $\frac{\exp(\mu_if_j^T)}{\sum_{j^{\prime}=1}^N\exp(\mu_if_{j^{\prime}}^T)}$ normalizes the probabilities across the target prototypes for each source prototype. $\sum_{i=1}^Np(f_i)$ ensures that the class distribution is properly taken into account when calculating the distillation cost. Then, minimizing $\mathcal{L}_{sep}$ encourages making more balanced mapping decisions, reducing the point-to-point mismatch risks, and thus promoting the separability of inter-class prototypes.

Finally, combining the compactness loss $\mathcal{L}_{com}$ and separability loss $\mathcal{L}_{sep}$, our point-to-point prototype topology distillation loss $\mathcal{L}_{ptd}$ is expressed as:
\begin{equation}
\mathcal{L}_{ptd}=\mathcal{L}_{com}+\mathcal{L}_{sep}.
\end{equation}
$\mathcal{L}_{ptd}$ mitigates the decision boundary overfitting to new classes at each session, thereby reducing the cost of maintaining old-class knowledge on the later adaptations.

\label{sec:ProposedMethod}

\section{Experiment}
\begin{table*}[!htbp]
\centering
\caption{Final accuracy~(\%) comparisons in the class-incremental scenario on Office-31-CI and ImageNet-Caltech-CI. U, CI, and SF respectively represent unsupervised, class-incremental, and source data-free. -B indicates the backbone is changed to ViT-B. * indicates the result is derived from ProCA.}
\resizebox{\textwidth}{!}{
\fontsize{9}{10}\selectfont
\begin{tabular}{@{}l@{\hskip 24pt}c@{\hskip 15pt}c@{\hskip 15pt}c@{\hskip 12pt}|c@{\hskip 24pt}c@{\hskip 24pt}c@{\hskip 24pt}c@{\hskip 24pt}c@{\hskip 24pt}c@{\hskip 24pt}c@{\hskip 15pt}|c@{\hskip 24pt}c@{\hskip 24pt}c@{}}
\toprule
\multirow{2}{*}{Method} & \multirow{2}{*}{U} & \multirow{2}{*}{CI} & \multirow{2}{*}{SF} & \multicolumn{7}{c|}{Office-31-CI} & \multicolumn{3}{c}{ImageNet-Caltech-CI} \\
\cmidrule(l{2pt}r{2pt}){5-11} \cmidrule(l{2pt}r{2pt}){12-14}
 & & & & A$\rightarrow$D & A$\rightarrow$W & D$\rightarrow$A & D$\rightarrow$W & W$\rightarrow$A & W$\rightarrow$D & Avg. & C$\rightarrow$I & I$\rightarrow$C & Avg. \\
\midrule
ViT-B & \xmark & \xmark & \xmark & 82.4 & 80.0 & 70.8 & 83.1 & 75.1 & 87.4 & 79.8 & 82.7 & 78.6 & 80.7 \\
CIDA*~(ECCV20) & \xmark & \cmark & \cmark & 70.4 & 64.5 & 48.1 & 95.1 & 52.7 & 98.8 & 71.6 & 69.3 & 49.2 & 59.2 \\
ProCA-B~(ECCV22) & \cmark & \cmark & \xmark & 93.4 & 91.9 & 73.0 & 98.3 & 77.6 & 99.2 & 88.9 & 91.6 & 84.0 & 87.8 \\
PLUE~(CVPR23) & \cmark & \xmark & \cmark & 74.5 & 74.6 & 70.3 & 85.7 & 70.5 & 80.1 & 76.0 & 82.4 & 70.6 & 76.5 \\
TPDS~(IJCV23) & \cmark & \xmark & \cmark & 78.1 & 74.6 & 71.1 & 90.3 & 69.7 & 91.9 & 79.3 & 85.0 & 68.0 & 76.5 \\
LCFD~(Arxiv24) & \cmark & \xmark & \cmark & 51.6 & 53.9 & 48.3 & 61.2 & 44.4 & 86.8 & 57.7 & 77.4 & 66.9 & 72.2 \\
DIFO~(CVPR24) & \cmark & \xmark & \cmark & 83.4 & 80.2 & 66.9 & 91.7 & 68.1 & 91.5 & 80.3 & 62.2 & 60.6 & 61.4 \\
LEAD-B~(CVPR24) & \cmark & \xmark & \cmark & 92.3 & 92.2 & 76.8 & 98.1 & 76.3 & \textbf{100.0} & 89.3 & 89.2 & 53.0 & 71.1 \\
\midrule
GROTO~(Ours) & \cmark & \cmark & \cmark & \textbf{99.4} & \textbf{99.0} & \textbf{81.3} & \textbf{99.0} & \textbf{81.3} & 98.1 & \textbf{93.0} & \textbf{92.5} & \textbf{85.1} & \textbf{88.8} \\
\bottomrule
\end{tabular}
\setlength{\tabcolsep}{6pt} 
}
\label{Table 1}
\end{table*}

\begin{table*}[!htbp]
\centering
\vspace{-5pt}
\caption{Final accuracy~(\%) comparisons in the class-incremental scenario on Office-Home-CI. 
}
\resizebox{\textwidth}{!}{
\fontsize{9}{10}\selectfont
\begin{tabular}{@{}lccc|ccccccccccccc@{}}
\toprule
Method & U & CI & SF & A$\rightarrow$C & A$\rightarrow$P & A$\rightarrow$R & C$\rightarrow$A & C$\rightarrow$P & C$\rightarrow$R & P$\rightarrow$A & P$\rightarrow$C & P$\rightarrow$R & R$\rightarrow$A & R$\rightarrow$C & R$\rightarrow$P & Avg. \\
\midrule
ViT-B  & \xmark  & \xmark & \xmark & 53.2 & 77.7 & 82.1 & 69.1 & 76.6 & 78.7 & 67.8 & 50.8 & 82.1 & 73.0 & 50.2 & 81.8 & 70.3 \\
CIDA*~(ECCV20)  & \xmark & \cmark & \cmark & 32.2 & 45.9 & 49.1 & 36.5 & 48.6 & 46.6 & 51.6 & 33.5 & 59.0 & 64.0 & 38.0 & 65.1 & 47.5 \\
ProCA-B~(ECCV22) & \cmark  & \cmark & \xmark & 60.5 & \textbf{88.0} & \textbf{92.9} & 83.3 & \textbf{89.7} & \textbf{91.0} & 81.7 & 57.1 & \textbf{94.1} & 86.9 & 55.8 & \textbf{92.9} & 81.2 \\
PLUE~(CVPR23) & \cmark & \xmark & \cmark & 28.8 & 67.9 & 72.5 & 60.5 & 67.5 & 73.4 & 59.6 & 29.2 & 74.2 & 61.2 & 37.6 & 72.6 & 58.8 \\
TPDS~(IJCV23) & \cmark & \xmark & \cmark & 40.5 & 63.5 & 69.0 & 64.7 & 67.3 & 68.7 & 63.8 & 40.8 & 71.5 & 68.2 & 29.4 & 65.8 & 59.4 \\
DIFO~(CVPR24) & \cmark & \xmark & \cmark & 44.7 & 62.6 & 61.5 & 54.1 & 59.8 & 61.6 & 53.7 & 37.6 & 65.2 & 58.4 & 43.3 & 57.5 & 55.0 \\
LEAD-B~(CVPR24) & \cmark & \xmark & \cmark & 33.9 & 81.8 & 86.9 & 76.1 & 82.4 & 84.8 & 74.0 & 19.2 & 87.3 & 79.7 & 15.2 & 84.3 & 67.1 \\
LCFD~(Arxiv24) & \cmark & \xmark & \cmark & 54.7 & 72.2 & 79.6 & 67.7 & 72.3 & 76.6 & 67.2 & 52.6 & 79.6 & 71.5 & 55.3 & 76.3 & 68.8 \\
\midrule
GROTO~(Ours) & \cmark & \cmark & \cmark & \textbf{65.7} & 86.4 & 89.7 & \textbf{85.8} & 86.3 & 90.0 & \textbf{86.0} & \textbf{67.1} & 90.1 & \textbf{86.9} & \textbf{66.2} & 89.3 & \textbf{82.5} \\
\bottomrule
\end{tabular}
}
\label{Table 2}
\vspace{-15pt}
\end{table*}


\subsection{Experimental Setup}
\noindent\textbf{Dataset.} We conduct experiments on the variants of three benchmark datasets, \textit{i.e.}, Office-31~\cite{saenko2010adapting}, Office-Home~\cite{venkateswara2017deep} and ImageNet-Caltech~\cite{russakovsky2015imagenet, griffin2020caltech}. 1) \textbf{\textit{Office-31-CI}} contains three domains with shared 31 classes in the office environment, \textit{i.e.}, Amazon~(A), DSLR~(D), and Webcam~(W). Each domain is divided into three disjoint subsets with 10 classes for each in alphabetical order. 2) \textbf{\textit{Office-Home-CI}} divides images of everyday objects into four domains with shared 65 classes, \textit{i.e}, Artistic~(A), ClipArt~(C), Product~(P), and Real-world~(R). Each domain is divided into six disjoint subsets with 10 classes for each in random order. 3) \textbf{\textit{ImageNet-Caltech-CI}} contains ImageNet-1K and Caltech-256 with shared 84 classes. It contains the ImageNet~(1000) $\rightarrow$ Caltech~(84) and Caltech~(256) $\rightarrow$ ImageNet~(84) tasks, where eight disjoint subsets of the target domain are constructed to contain 10 classes for each. Refer to the supplementary material for more dataset construction details.

\noindent\textbf{Implementation details.} 
We utilize SGD as the optimizer and the base learning rate is set to 0.001 with the batch size of 32. 
The number of incremental classes $\gamma$ at each session is set to 10. In the early iterations, we identify multi-granularity class prototypes with the source model and then use the well-trained target model. For $\mathcal{L}_{con}$, we exponentially decrease the coefficient $\mu_{c}$ as $\mu_c^i=\mu_c^{i-1}e^{-\beta}$ at the $i$-th iteration, where initial $\mu_{c}^0$ is set to 0.5 and ${\beta}$ is set to 1e-4. Moreover, the stored exemplar $n_r$ of each existing positive class in the memory bank is set to 10.

\noindent\textbf{Evaluation protocols.} 1) \textbf{\textit{Final Accuracy:}} the classification accuracy over all the classes at the final session. 
2) \textbf{\textit{Session Accuracy:}} the classification accuracy at each session to evaluate the ability of sequential adaptation.

\begin{table*}[!htbp]
\centering
\caption{Session accuracy~(\%) comparisons in the class-incremental scenario on Office-31-CI and Office-Home-CI. 
}
{\fontsize{9}{10}\selectfont
\resizebox{\textwidth}{!}{
\fontsize{9}{10}\selectfont
\begin{tabular}{l@{\hskip 20pt}c@{\hskip 10pt}c@{\hskip 10pt}c@{\hskip 10pt}|cccc|ccccccc}
\toprule
\multirow{2}{*}{Method} & \multirow{2}{*}{U} & \multirow{2}{*}{CI} & \multirow{2}{*}{SF} & \multicolumn{4}{c|}{Office-31-CI} & \multicolumn{7}{c}{Office-Home-CI} \\
\cmidrule(lr){5-8} \cmidrule(l){9-15}
 & & & & Session 1 & Session 2 & Session 3 & Avg. & Session 1 & Session 2 & Session 3 & Session 4 & Session 5 & Session 6 & Avg. \\
\midrule
CIDA*~(ECCV20) & \xmark & \cmark & \cmark & 85.5 & 79.1 & 71.6 & 78.7 & 57.9 & 53.6 & 51.8 & 50.1 & 49.6 & 47.5 & 51.8 \\
ProCA-B~(ECCV22) & \cmark  & \cmark & \xmark & 90.5 & 90.0 & 89.0 & 89.8 & 76.3 & 81.0 & 80.1 & 81.0 & 81.6 & 81.2 & 80.2 \\
DIFO~(CVPR24) & \cmark & \xmark & \cmark & 82.0 & 66.8 & 80.3 & 76.4 & 77.2 & 37.1 & 47.9 & 51.4 & 51.7 & 55.0 & 53.4 \\
LCFD~(Arxiv24) & \cmark & \xmark & \cmark & 81.3 & 68.9 & 57.7 & 69.3 & 70.0 & 64.9 & 68.2 & 68.1 & 67.9 & 68.8 & 68.0 \\
LEAD-B~(CVPR24) & \cmark & \xmark & \cmark & 96.4 & 91.6 & 89.3 & 92.4 & 70.5 & 74.0 & 71.2 & 70.5 & 69.0 & 67.1 & 70.4 \\
\midrule
GROTO~(Ours) & \cmark & \cmark & \cmark & \textbf{96.7} & \textbf{94.9} & \textbf{93.0} & \textbf{94.9} & \textbf{84.4} & \textbf{84.8} & \textbf{81.7} & \textbf{83.3} & \textbf{82.8} & \textbf{82.5} & \textbf{83.2} \\
\bottomrule
\label{Table 3}
\end{tabular}
}}
\vspace{-15pt}
\end{table*}

\subsection{Comparisons with SOTA Methods}
We compare our GROTO algorithm with five types of methods: (1) source-only: ViT-B~\cite{dosovitskiy2020image}; (2) class-incremental domain adaptation: CIDA~\cite{kundu2020class}; (3) class-incremental unsupervised domain adaptation: ProCA~\cite{lin2022prototype}; (4) source-free unsupervised domain adaptation: PLUE~\cite{litrico2023guiding} and TPDS~\cite{tang2024source}; (5) source-free universal domain adaptation: DIFO~\cite{tang2024sourcefreedomainadaptationfrozen}, LEAD~\cite{qu2024lead} and LCFD~\cite{tang2024unified}.

The final accuracy results of GROTO and the compared methods are reported in ~\Cref{Table 1} and ~\Cref{Table 2}, which give the following observations. 
1) GROTO shows the superiority of average final accuracy in all sub-tasks compared to other methods. 
2) Compared to SFUDA methods~\cite{litrico2023guiding, tang2024source}, GROTO performs better. It demonstrates the importance of reducing negative transfer caused by source negative classes and adapting to new classes while retaining old knowledge in CI-SFUDA. 3) The CIDA~\cite{kundu2020class} method uses the regularization on proxy-source samples to mitigate catastrophic forgetting in source-free domain adaptation.
However, it overlooks the target-class biased learning caused by the source negative classes, limiting its effectiveness in CI-SFUDA. 4) The CI-UDA method~(ProCA~\cite{lin2022prototype}) preserves old knowledge through the exemplar replay method. However, it relies on source data, which makes it difficult to satisfy data privacy issues. 5) SF-UniDA methods~\cite{tang2024sourcefreedomainadaptationfrozen,qu2024lead, tang2024unified} effectively prevent target features from shifting toward similar source clusters outside the target label space. However, they perform even worse than ViT-B in some sub-tasks, which shows that only focusing on domain alignment without preventing the forgetting leads the model to disrupt old knowledge and overfit the new classes in CI-SFUDA.

\begin{table}[t]
\centering
\vspace{-8pt}
\caption{Ablations of losses~(\textit{i.e.}, $\mathcal{L}_{rep}$, $\mathcal{L}_{ptfs}$, $\mathcal{L}_{ptd}$,$\mathcal{L}_{con}$,$\mathcal{L}_{com}$ and $\mathcal{L}_{sep}$), components~(\textit{i.e.}, coarse-grained $O_c$ and fine-grained prototypes $O_f$) and modules~(\textit{i.e.}, HKPCM, PTFS and PTD) in GROTO. We show the final accuracy~(\%) on Office-31-CI.}
\fontsize{9}{10}\selectfont
\setlength{\tabcolsep}{0.3pt}
\begin{tabular}{ccccccccccc}
\toprule
\multirow{2}{*}{Exps.} & \multirow{2}{*}{$\mathcal{L}_{rep}$} & \multirow{2}{*}{$\mathcal{L}_{ptfs}$} & \multirow{2}{*}{$\mathcal{L}_{ptd}$} & \multicolumn{6}{c}{Final Accuracy~(\%)} \\ 
\cmidrule(l){5-11}
 & & & & A$\rightarrow$D & A$\rightarrow$W & D$\rightarrow$A & D$\rightarrow$W & W$\rightarrow$A & W$\rightarrow$D & Avg. \\ 
\midrule
A &  &  &  & 82.4 & 80.0 & 70.8 & 83.1 & 75.1 & 87.4 & 79.8 \\
B & \cmark &  &  & 90.7 & 91.9 & 74.0 & 98.1 & 77.2 & 98.3 & 88.4 \\
C & \cmark & \cmark &  & 96.5 & 97.3 & 80.9 & 99.2 & 80.6 & 97.5 & 92.0 \\
D & \cmark & \cmark & \cmark & \textbf{99.4} & \textbf{99.0} & \textbf{81.3} & 99.0 & \textbf{81.3} & 98.1 & \textbf{93.0} \\
\midrule[\heavyrulewidth]
\multicolumn{4}{l}{Exps.B w/ $O_f$} & 93.8 & 96.4 & 78.6 & 97.9 & 80.1 & 97.3 & 90.7 \\
\multicolumn{4}{l}{Exps.B w/ $O_c$} & 95.2 & 97.4 & 78.2 & 98.2 & 80.2 & 98.8 & 91.3 \\
\multicolumn{4}{l}{Exps.C w/o $\mathcal{L}_{con}$} & 94.4 & 97.3 & 79.0 & 99.1 & 80.7 & 98.6 & 91.5 \\
\multicolumn{4}{l}{GROTO w/o $\mathcal{L}_{com}$} & 95.7 & 97.3 & 80.6 & 99.2 & 81.1 & 99.6 & 92.3 \\
\multicolumn{4}{l}{GROTO w/o $\mathcal{L}_{sep}$} & 95.4 & 99.1 & 80.3 & 99.0 & 80.3 & 100.0 & 92.4 \\
\multicolumn{4}{l}{GROTO w/o PTD} & 96.5 & 97.3 & 80.9 & 99.2 & 80.6 & 97.5 & 92.0 \\
\multicolumn{4}{l}{GROTO w/o PTFS} & 88.4 & 91.6 & 73.8 & 98.2 & 77.3 & 99.0 & 88.0 \\
\multicolumn{4}{l}{GROTO w/o HKPCM} & 97.5 & 95.5 & 80.6 & 99.2 & 75.1 & 98.1 & 91.0 \\
\bottomrule
\end{tabular}
\label{Table 4}
\vspace{-15pt}
\end{table}

\begin{figure}
\centering
\vspace{-5pt}
\includegraphics[width=1.0\columnwidth]{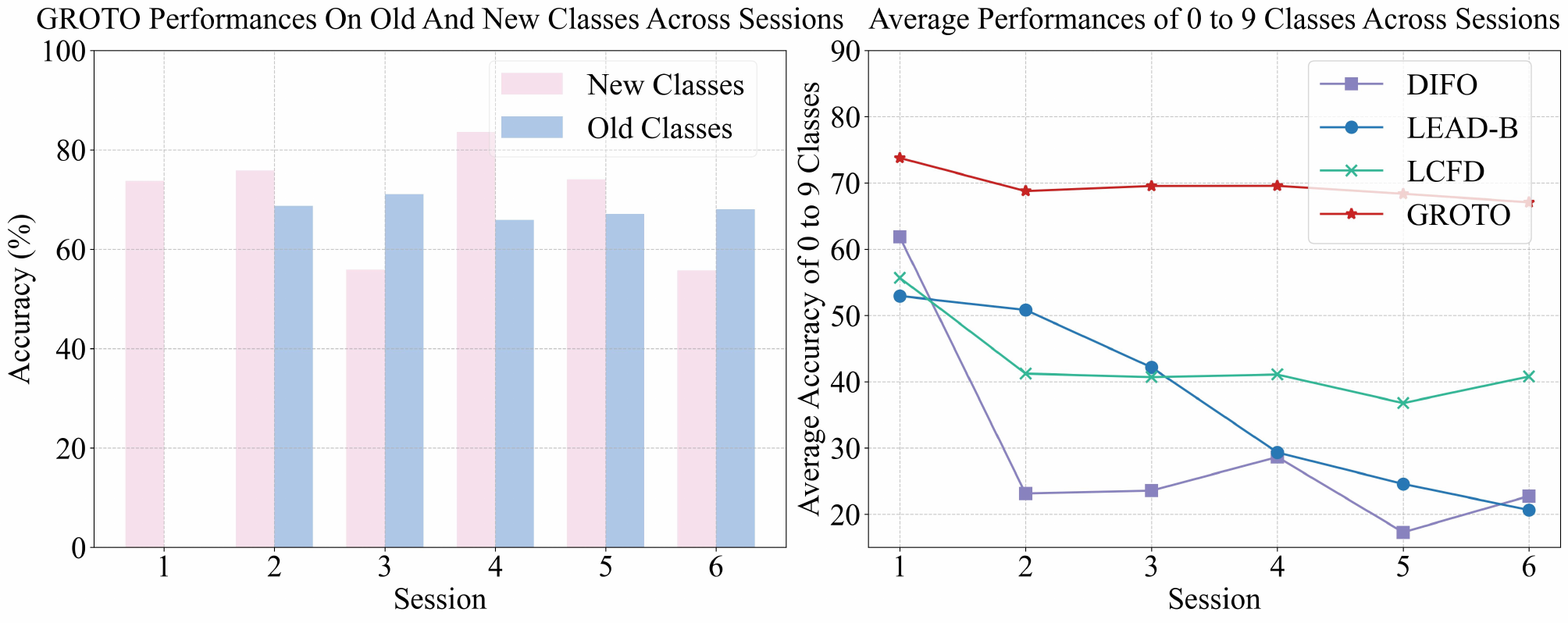} 
\caption{The accuracies on old and new classes of GROTO on Office-Home-CI~(R$\rightarrow$C), and the average accuracies of 0 to 9 classes for different methods across sessions on Office-Home-CI~(R$\rightarrow$C).}
\label{Figure 3}
\vspace{-15pt}
\end{figure}

The session accuracy comparisons are shown in ~\Cref{Table 3}. There is no catastrophic forgetting problem at session 1, thus the SF-UniDA methods perform well. The best performance of our GROTO at session 1, illustrating the effectiveness of the multi-granularity class prototype self-organization module. Meanwhile, our GROTO achieves the best performance compared to the other methods at each session in ~\Cref{Table 3}, which also demonstrates that it performs well on both old and new classes during the continual adaptation of sequential sessions. ~\Cref{Figure 3} further explores the ability to maintain the performance of old classes. Since the absence of old-class instances at later sessions, the models of all the methods forget the old knowledge when adapting to new classes, resulting in performance degradation. But other methods continually show significant drops in the average accuracy of 10 classes learned at session 1, our GROTO maintains relatively stable performances compared to them, indicating that it is effective in mitigating the catastrophic forgetting problem. Refer to the supplementary material for more dataset results and analysis.

\subsection{Ablation Studies}
\noindent\textbf{Loss and Module Ablations.} To examine the effectiveness of our GROTO, we show the ablations in ~\Cref{Table 4}. Compared to the source-only model~(Exps.A), the final accuracy is improved with the introduction of each loss, and it decreases when each proposed module is removed. 

\begin{figure}
\centering
\includegraphics[width=1\columnwidth]{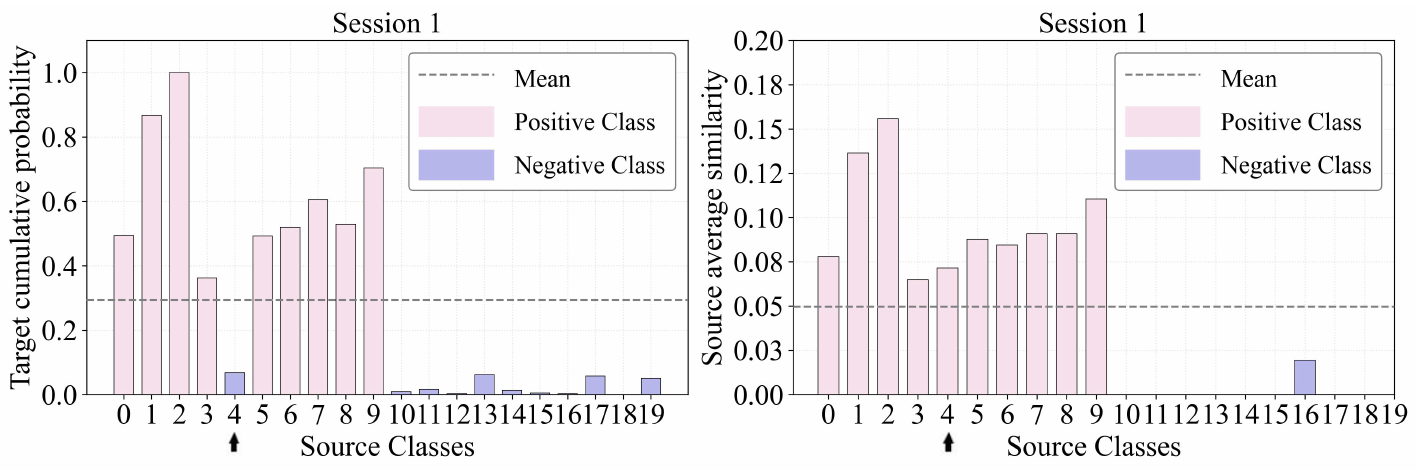}
\caption{The target cumulative probability and source average similarity results at the first session on Office-31-CI~(A$\rightarrow$D). Only training data from 0 to 9 classes are provided at this session. Based on both distributions, we select classes with indices 0-9 as the positive classes at this session.}
\label{Figure 4}
\vspace{-15pt}
\end{figure}

Target probability and source similarity accumulation distributions of Office-31-CI~(A$\rightarrow$D) at session 1 are shown in ~\Cref{Figure 4}. We can observe that both the positive-class ones are higher than the negative-class ones. In the left sub-figure, only depending on the model predictions to mine the positive classes leads to omissions of the positive class ``4"~(pointed by the black arrow), resulting in training with the wrong pseudo-labels at new sessions. Thus, the HKPCM module is proposed to prevent the missing positive classes due to the adoption of any single cumulative distribution strategy, and its effectiveness is verified through the ablation experiment of ``GROTO w/o HKPCM” in ~\Cref{Table 4}.


\begin{figure}
\centering
\includegraphics[width=1\columnwidth]{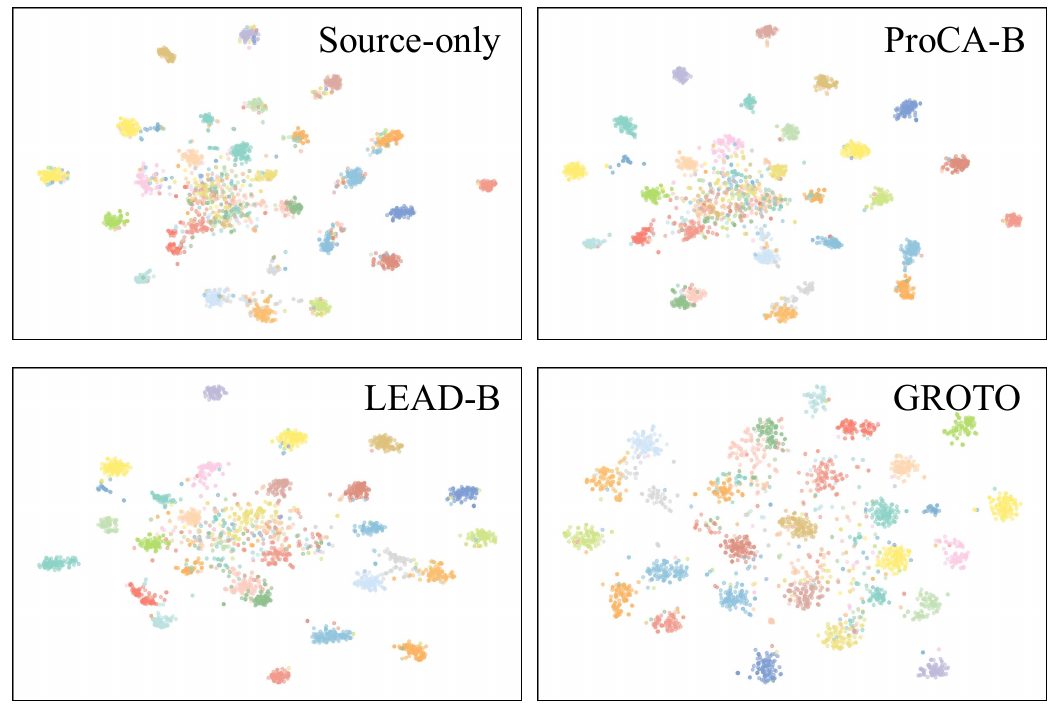} 
\caption{The multi-granularity class prototype visualizations of partial classes on Office-31-CI~(A$\rightarrow$D), and target feature distribution of source-only, ProCA-B, LEAD-B and GROTO methods on Office-31-CI~(D$\rightarrow$A).}
\label{Figure 5}
\end{figure}

\begin{figure}
\centering
\vspace{-10pt}
\includegraphics[width=1\columnwidth]{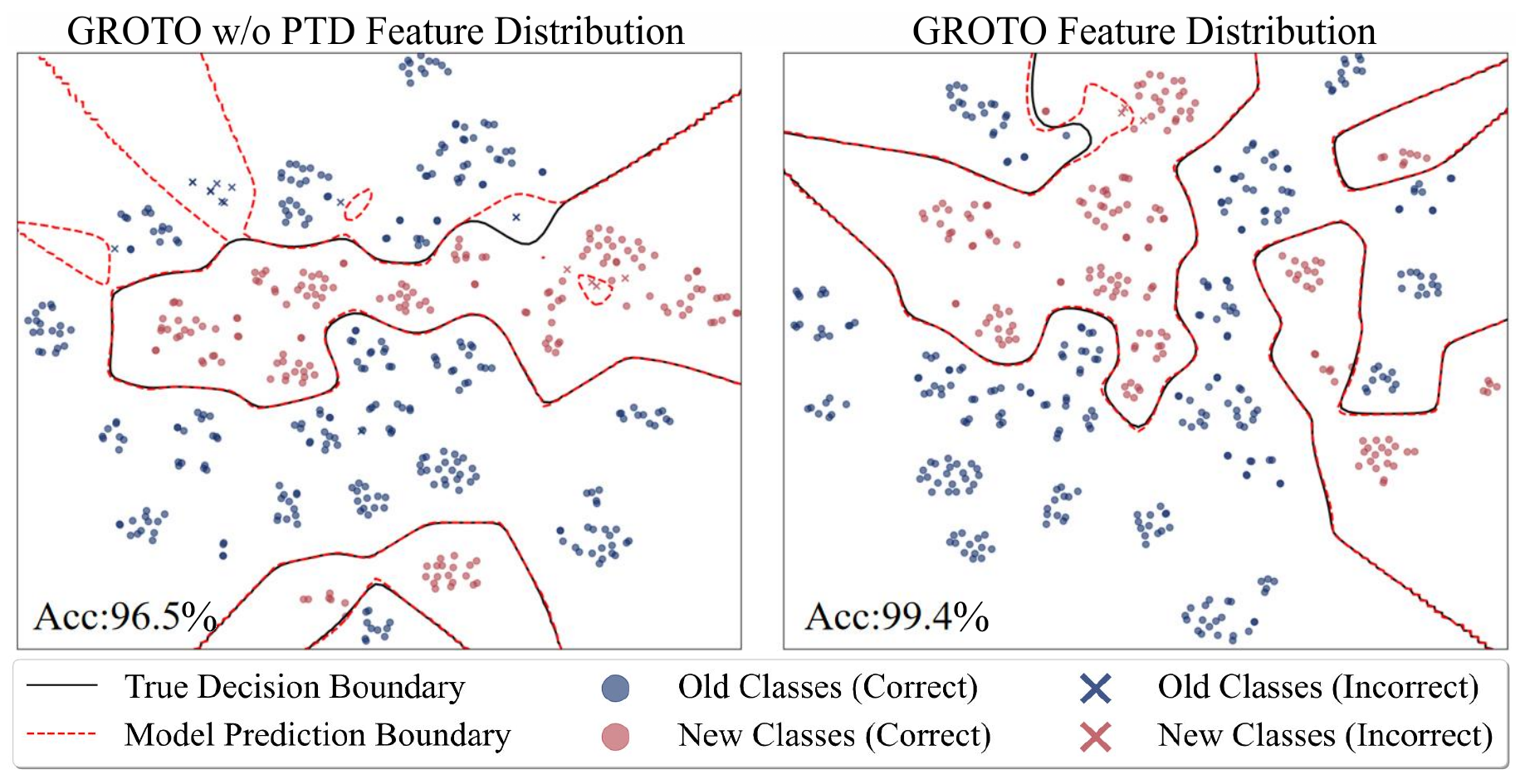} 
\caption{The feature distribution and decision boundary visualizations of ``GROTO w/o PTD" and GROTO models on Office-31-CI~(A$\rightarrow$D). ``GROTO w/o PTD" tends to incorrectly classify old-class data to new ones.}
\label{Figure 6}
\vspace{-15pt}
\end{figure}

The Exps.B experiment introduces the HKPCM module and $\mathcal{L}_{rep}$, but it relies on simple model predictions to generate pseudo-labels for unlabeled target data. In the Exps.C experiment of introducing the $\mathcal{L}_{rep}$, HKPCM and PTFS modules, we first mine positive classes with the HKPCM module, and then generate pseudo-labels with the PTFS module. Compared with the Exps.B experiment, the performance is improved through the introduction of the PTFS module. It can also be seen that the combination of coarse-grained and fine-grained prototypes demonstrates superior performance compared to using either prototype type individually in ~\Cref{Table 4}. When we introduce $\mathcal{L}_{con}$, the performance is further improved. In addition, we show the target feature distributions of different methods on Office-31-CI~(D$\rightarrow$A) in~\Cref{Figure 5}. The source-only model~(ViT-B), ProCA-B, and LEAD-B mix the target features of similar classes. However, our GROTO alleviates the problem, showing a more discriminative cluster distribution among different classes.



The experiments of Exps.D, ``GROTO w/o $\mathcal{L}_{com}$", ``GROTO w/o $\mathcal{L}_{sep}$" and ``GROTO w/o PTD" demonstrate that the PTD module improves the final accuracy. To investigate its impact on old-class knowledge, we visualize old-class and new-class features at the last session on Office-31-CI~(A$\rightarrow$D) as shown in ~\Cref{Figure 6}. GROTO without the PTD module~(left sub-figure) incorrectly classifies old-class data to new ones in the top left part. However, fewer old data are misclassified to new classes with the PTD module, thus improving the overall classification accuracy of old and new classes~(96.5\%$\rightarrow$99.4\%). It also proves that the PTD module effectively mitigates the problem of excessively focusing on new classes while forgetting old knowledge on continual adaptation. 

\begin{table}[t]
\centering  
\caption{Ablations of different hyper-parameters~(\textit{i.e.},$\gamma$, $\mu_c^0$, $\beta$ and $n_r$) for the average final accuracy~(\%) on ImageNet-Caltech-CI.}  
\fontsize{9}{10}\selectfont
\begin{tabular}{cc|cc|cc|cc}         
\toprule
$\gamma$ & Avg.  & $\mu_c^0$ & Avg.  & $\beta$  & Avg. & $n_r$ & Avg. \\
\midrule
9  & 87.5 & 0.2 & 88.5 & 1e-5 & 88.7 & 1 & 83.0 \\
10  & 88.8 & 0.5 & \textbf{88.8} & 5e-5 & 88.8 & 5 & 86.2 \\
11 & 87.8 & 0.6 & 87.5 & 1e-4 & \textbf{88.8} & 10 & \textbf{88.8} \\
12 & 87.3 & 0.7 & 85.4 & 5e-4 & 87.3 & 15 & 88.7 \\
13 & \textbf{89.1} & 0.8 & 88.0 & 1e-3 & 88.3 & 20 & 88.7 \\
\bottomrule
\end{tabular}
\vspace{-10pt}
\label{Table 5}
\end{table}

\noindent\textbf{Hyper-parameter Ablations.} We explore the impacts of hyper-parameters in ~\Cref{Table 5}, indicating that GROTO has relatively stable performance with different values of $\mu_c^0$ and $\beta$. The final accuracy is highest when the number of incremental classes $\gamma$ is 13, but to ensure the experimental setup is consistent with ProCA, we choose 10 as the number of
incremental classes for GROTO. 

\label{sec:Experiment}
\section{Conclusion}
In this paper, we explore the class-incremental source-free unsupervised domain adaptation~(CI-SFUDA) problem. To address it, we propose a novel Multi-Granularity Class Prototype Topology Distillation~(GROTO) algorithm, which contains two modules. (1) Multi-granularity class prototype self-organization: we mine positive classes to align the label spaces. Next, we introduce multi-granularity class prototypes to generate reliable pseudo-labels, and promote the target feature self-organization.
(2) Prototype topology distillation: we construct the topological structures of feature manifolds with source and target prototypes, and perform the point-to-point topology distillation to reduce the decision boundary overfitting to new target classes. Therefore, we can continually mitigate the shocks of new target knowledge to old ones. Extensive experiments demonstrate the effectiveness of our GROTO on three public datasets, and it achieves state-of-the-art performance.

\label{sec:conclusion}

\section*{Acknowledgement}
 \begin{sloppypar}
This work was supported by the National Nature Science Foundation of China~(U21B2024, 62322211, 62336008), the ``Pioneer" and ``Leading Goose" R\&D Program of Zhejiang Province~(2024C01107, 2023C01030, 2023C03012, 2024C01023).
\end{sloppypar}


\clearpage
\setcounter{page}{1}
\maketitlesupplementary

\section{Overview}
In the supplementary material, we provide more details and analysis about the algorithm. We also provide more results and visualizations to demonstrate the effectiveness of GROTO. The materials are organized as follows.

\noindent1) In~\Cref{Section 1}, we provide pseudo-codes of the training pipeline and each module.

\noindent2) In~\Cref{Section 3}, we provide more results of the large-scale dataset.

\noindent3) In~\Cref{Section 4}, we provide the effective analysis of the hybrid knowledge-driven positive class mining~(HKPCM) module.

\noindent4) In~\Cref{Section 5}, we provide more analysis of the prototype topology distillation~(PTD) module.

\noindent5) In~\Cref{Section 6}, we provide session accuracy comparisons.

\noindent6) In~\Cref{Section 7}, we provide more details on the training process of GROTO.

\noindent7) In~\Cref{Section 8}, we provide more visualizations of different methods on Office-31-CI.

\noindent8) In~\Cref{Section 9}, we provide more details on constructing the incremental datasets.

\noindent9) In Section~\ref{section3}, we provide training cost and the complexity analysis of our GROTO algorithm.

\section{More Pseudo-codes of the algorithm}
In this section, we provide the pseudo-codes of the training process of GROTO, the hybrid knowledge-driven positive class mining~(HKPCM), the positive-class target feature self-organization~(PTFS), the prototype topology distillation~(PTD), and the knowledge replay modules~(KR), respectively.
\label{Section 1}

\begin{algorithm}
\caption{Training process of GROTO}
\begin{algorithmic}[1]
\State \textbf{Input:} The target model $M_t$ and the training epoch $E$
\State Mine the positive classes through the HKPCM module
\For{$i = 1 \rightarrow E$}
    \State Assign pseudo-labels of the unlabeled positive-class data
    \State Compute $\mathcal{L}_{ce}$ and $\mathcal{L}_{con}$
    \State Compute $\mathcal{L}_{ptd}$
    \State Compute $\mathcal{L}_{rep}$
    \State Update the target model $M_t$ by optimizing
    \State \(\min \left( \mathcal{L}_{ce} + \mu_c^i\mathcal{L}_{con} + \mathcal{L}_{ptd} + \mathcal{L}_{rep} \right)\)
\EndFor
\State \textbf{return} the target model $M_t$
\label{Algorithm1}
\end{algorithmic}
\end{algorithm}

\begin{algorithm}
\caption{Hybrid Knowledge-driven Positive Class Mining}
\begin{algorithmic}[2]
\State \textbf{Input:} Source feature centroids $s=[s_1, ..., s_K]$ of all the source classes $K$, the target data $x=[x_1, ..., x_{n_t}]$ of the $t$-th session, target features $g=[g_1, ..., g_{n_t}]$ of the $t$-th session, the source feature extractor $G_s$ and the source classifier $C_s$
\State \textbf{Output:} Mined Positive Classes
\State \code{/*Compute the similarity matrix*/} 
\State $R = g^T \cdot s$, where each element $r_{ij} \in {R}$ represents the similarity between the $i$-th target data and the ${j}$-th source centroid
\For{$i = 1 \rightarrow n_t$}
    \For{$j = 1 \rightarrow K$}
        \State \code{/*Compute the softmax normalization*/}
        \State $S_{ij}=\frac{e^{r_{ij}}}{\sum_{k=1}^K e^{r_{ik}}}$
    \EndFor
\EndFor
\State \code{/*Compute the average similarity probability*/}
\For{$k = 1 \rightarrow K$}
    \State $S_k = \frac{1}{n_t} \sum_{i=1}^{n_t} S_{ik}$
\EndFor

\State \code{/*Identify the positive classes*/}
\For{$k = 1 \rightarrow K$}
    \If{$S_k > \frac{1}{K}\sum_{j=1}^{K}S_j$}
        \State Class $k$ is a positive class
    \EndIf
\EndFor

\For{$k = 1 \rightarrow K$}
   \State \code{/*Compute the cumulative probability*/} 
   \State $P_k = \sum_{i=1}^{n_t} C_s(G_s(x_i))$
\EndFor

\State $P =[P_1, P_2, ..., P_K]$, where $P$ is the probabilities of all the source classes $K$ 

\For{$k = 1 \rightarrow K$}
 \State \code{/*Normalize the probability*/}
 \State $P_k = \frac{P_k - \min(P)}{\max(P) - \min(P)}$
\EndFor
\State \code{/*Identify the positive classes*/} 
\For{$k = 1 \rightarrow K$}
    \If{$P_k > \frac{1}{K} \sum_{j=1}^{K} P_j$}
        \State Class $k$ is a positive class
    \EndIf
\EndFor

\label{Algorithm 2}
\end{algorithmic}
\end{algorithm}

\begin{algorithm}
\caption{Positive-class Target Feature Self-organization}
\begin{algorithmic}[3]
\State \textbf{Input:} Target data $x=[x_1, ..., x_{n_t}]$ and their weak augmented data $x^{'}=[x_1^{'}, ..., x_{n_t}^{'}]$ of the $t$-th session, the target feature extractor $G_t$, the target classifier $C_t$, and source classifier weights $\mu=[\mu_1,...,\mu_{N}]$ of $N$ positive classes
\State \textbf{Output:} Pseudo-labels, the loss of PTFS module $\mathcal{L}_{ptfs}$
\State \code{/*Identify source coarse-grained prototypes*/} $\mu{\in}O_c$

\For{$i = 1 \rightarrow n_t$}
    \State $\bar{y}_i=\arg\max_n{C_t(G_t(x_i))}$
\EndFor

\For{$n = 1 \rightarrow N$}
    \State \code{/*$S_n$ is the set of data that initial pseudo-label is $n$*/} 
    \State $c_n=\frac{1}{|S_n|} \sum_{x_n \in S_n} G_t(x_n)$
    \For{each $x_n \in S_n$}
        \State $d(x_n,c_n)=1-\frac{G_t(x_n)\cdot c_n}{\|G_t(x_n)\|_2 \|c_n\|_2}$
    \EndFor
    \State $\tau_s=\frac1{\mid S_n\mid}\sum_{x_n\in S_n}d(x_n,c_n)$
    \State \code{/*Identify the target coarse-grained prototypes*/}
    \For{each $x_n \in S_n$}
        \If{$d(x_n,c_n) < \tau_s$}
            \State $x_n{\in}O_c$
        \EndIf
    \EndFor
\EndFor

\For{$i = 1 \rightarrow n_t$}
    \State $u_i=std(conf(x_i), conf(x_i^{'}))$
\EndFor
\State Compute $\tau_c=\frac1{2n_t}\sum_{i=1}^{n_t}\left(conf(x_i)+conf(x_i^{'})\right)$
\State Compute $\tau_u=\frac1{n_t}\sum_{i=1}^{n_t}u_i$
\State \code{/*Identify the fine-grained prototypes*/}
\For{$i = 1 \rightarrow n_t$}
    \If{$conf_{avg}(x_i, x_i^{'}) > \tau_c$ and $u_i < \tau_u$}
        \State $x_i{\in}O_f$
    \EndIf
\EndFor

\State Compute $O=O_c\cup O_f$
\For{$j = 1 \rightarrow n_t-{\mid O \mid}$}
    \State \code{/*Compute the distance to all $n$-class prototypes*/}
    \For {$i = 1 \rightarrow \mid o_n \mid$}
        \State $d(x_j, o_{i,n})=1 - \frac{G_t(x_j) \cdot o_{i,n}}{\|G_t(x_j)\|_2 \|o_{i,n}\|_2}$
    \EndFor
    \State \code{/*The distance of $x_j$ to the $n$-th positive-class*/}
    \State $D(x_j,n)=\frac{1}{\mid o_n\mid}\sum_{o_{i,n} \in o_n}d(x_j,o_{i,n})$
    \State $\bar{y}_j=\arg\min_n{D(x_j,n)}$
\EndFor

\State Compute $\mathcal{L}_{ce}=-\frac{1}{B}\sum_{i=1}^{B}\bar{y}_{i}\cdot\log \left(C_t(G_t(x_i)) \right)$
\State $\ell_{i,j}=-\log\frac{\exp(\operatorname\phi(G_t(x_i),G_t(x_i^{'}))/\kappa)}{\sum_{b=1}^{2B}\mathbbm{1}_{b\neq i}\exp(\operatorname\phi(G_t(x_i),G_t(x_i^{'}))/\kappa)}$
\State Compute $\mathcal{L}_{con}=\frac1{2B}\sum_{b=1}^{2B}[\ell_{2b-1,2b}+\ell_{2b,2b-1}]$
\State Compute $\mathcal{L}_{ptfs}=\mathcal{L}_{ce}+\mathcal{L}_{con}$
\label{Algorithm 3}
\end{algorithmic}
\end{algorithm}

\begin{algorithm}
\caption{Prototype Topology Distillation}
\begin{algorithmic}[4]
\State \textbf{Input:} The target class proportion $p(\cdot)$, the source classifier weights $\mu=[\mu_1,...,\mu_{N}]$ and target classifier weights $f=[f_1,...,f_{N}]$ of $N$ positive classes
\State \textbf{Output:} The prototype topology distillation loss $\mathcal{L}_{ptd}$

\State Initialize $\mathcal{L}_{com} = 0$, $\mathcal{L}_{sep} = 0$
\For{$j = 1 \rightarrow N$}
    \State \code{/*Initialize normalization factor*/} $Z = 0$
    \For{$i^{'} = 1 \rightarrow N$}
        \State $Z = Z + p(f_{i^{'}}) \exp(\mu_{i^{'}} f_j^T)$
    \EndFor
    \For{$i = 1 \rightarrow N$}
        \State \code{/*Cosine distance*/} $d(\mu_i, f_j^T)$
        \State $\mathcal{L}_{com}  \textbf{+=} \frac{1}{N} d(\mu_i, f_j^T) \frac{p(f_i) \exp(\mu_i f_j^T)}{Z}$
    \EndFor
\EndFor

\For{$i = 1 \rightarrow N$}
    \State \code{/*Initialize normalization factor*/} $Z' = 0$
    \For{$j^{'} = 1 \rightarrow N$}
        \State $Z' = Z' + \exp(\mu_i f_{j^{'}}^T)$
    \EndFor
    \For{$j = 1 \rightarrow N$}
        \State $\mathcal{L}_{sep} \textbf{+=} p(f_i) d(\mu_i, f_j^T) \frac{\exp(\mu_i f_j^T)}{Z'}$
    \EndFor
\EndFor
\State Compute $\mathcal{L}_{ptd} = \mathcal{L}_{com} + \mathcal{L}_{sep}$
\label{Algorithm 4}
\end{algorithmic}
\end{algorithm}

\begin{algorithm}
\caption{Knowledge Replay}
\begin{algorithmic}[5]
\State \textbf{Input:} The number of positive classes $N$ at the $t$-th session, all the target data $S_n$ of pseudo-label $n$, $x_n$ is each data in $S_n$, the target feature extractor $G_t$, the target classifier $C_t$, the number of all the stored existing positive classes in the memory bank $N^{'}$, the number of stored representative exemplars for each of the existing class $n_r$
\State \textbf{Output:} The exemplar replay loss $\mathcal{L}_{rep}$

\For{$n = 1 \rightarrow N$}
    \State \code{/*Compute centroids*/} $f_n = \frac{1}{\mid S_n \mid} \sum_{x_n \in S_n} G_t(x_n)$
    \For{$k = 1 \rightarrow n_r$}
    \Statex $m_{n}^{k}\!=\!{\arg\min}_{x_n \in S_n}\!\|f_n\!-\!\frac{1}{k}(G_t(x_n)\!+\!\sum_{i=1}^{k-1}\!G_t(m_i^n))\|_{2}$
    \State Store $m_{n}^{k}$ in the memory bank
    \EndFor
\EndFor

\State \code{/*Maintain the memory bank*/} $\mathcal{M}=\{(m_i,\hat{y}_i,\bar{y}_i)\}_{i=1}^{N_r}$
where $m_{i}$, $\hat{y}_i$, $\bar{y}_i$ and $N_r$ respectively denote the representative exemplar, soft prediction, pseudo-label, and the number of all the stored exemplars
\State Compute $\mathcal{L}_{rep} = -\frac{1}{N_r}\sum_{i=1}^{N_r} {\hat{y}_i}^\top \log C_t(G_t(m_i))$
\label{Algorithm 5}
\end{algorithmic}
\end{algorithm}

\section{More results of the large-scale dataset} 
\label{Section 3}
In this section, we add more experiments of the 12 sub-tasks on the large-scale DomainNet-126-CI dataset. DomainNet-126-CI divides images of common objects into four distinct domains with shared 126 classes, \textit{i.e}, Clipart~(C), Painting~(P), Sketch~(S), and Real~(R). Each domain is divided into six disjoint subsets with 20 classes for each in alphabetical order. 
When taking the C, P, S, and R as the target domain respectively, the average final accuracy across three sub-tasks for each target domain is shown in ~\Cref{Table 11}. Our GROTO outperforms DIFO~(+22.2\%) and LCFD~(+3.2\%), which demonstrates the superiority of our algorithm in handling the CI-SFUDA problem.

\begin{table}[!ht]
    \centering
    \caption{Average final accuracy~(\%) comparisons on 12 sub-tasks of DomainNet-126-CI.}
    \begin{adjustbox}{width=\linewidth,center}
    \begin{tabular}{lcccccc}
        \toprule
        Method &  $\rightarrow$C & $\rightarrow$P &  $\rightarrow$S &  $\rightarrow$R & $\rightarrow$Avg. \\
        \midrule
        DIFO~(CVPR24)  & 42.2  & 46.7  & 36    & 49.1  & 43.5  \\
        LCFD~(Arxiv24) & 60.1  & 61.4  &  \textbf{57.7}  & 70.8  & 62.5  \\
        GROTO~(Ours)   & \textbf{66.2}  & \textbf{61.7}  & 56.7  & \textbf{78.0}    & \textbf{65.7}  \\
        \bottomrule
    \end{tabular}
    \label{Table 11}
    \end{adjustbox}
\end{table}

\section{More results of the HKPCM module}
\label{Section 4}
\begin{table}[!htbp]
\centering
\footnotesize
\setlength{\tabcolsep}{10pt}
\caption{Mined accuracy (\%) comparisons of the HKPCM module with the ProCA-B method on ImageNet-Caltech-CI.}
\begin{tabular}{llcccc}
\toprule
Session & Method & Metric & C$\rightarrow$I & I$\rightarrow$C \\
\midrule
\multirow{6}{*}{Session 1} & \multirow{3}{*}{ProCA-B} 
    & PCD Acc. & 90 & 100 \\
    & & TCD Acc. & 69.2 & 100 \\
    & & Avg. & 79.6 & 100 \\
\cmidrule(lr){2-5}
    & \multirow{3}{*}{GROTO}
    & PCD Acc. & 90 & 100 \\
    & & TCD Acc. & 81.8 & 100 \\
    & & Avg. & \textbf{85.9} & \textbf{100} \\
\midrule
\multirow{6}{*}{Session 2} & \multirow{3}{*}{ProCA-B}
    & PCD Acc. & 100 & 90 \\
    & & TCD Acc. & 83.3 & 69.2 \\
    & & Avg. & 91.7 & 79.6 \\
\cmidrule(lr){2-5}
    & \multirow{3}{*}{GROTO}
    & PCD Acc. & 100 & 100 \\
    & & TCD Acc. & 100 & 76.9 \\
    & & Avg. & \textbf{100} & \textbf{88.5} \\
\midrule
\multirow{6}{*}{Session 3} & \multirow{3}{*}{ProCA-B}
    & PCD Acc. & 100 & 100 \\
    & & TCD Acc. & 100 & 100 \\
    & & Avg. & \textbf{100} & \textbf{100} \\
\cmidrule(lr){2-5}
    & \multirow{3}{*}{GROTO}
    & PCD Acc. & 100 & 100 \\
    & & TCD Acc. & 90.9 & 90.9 \\
    & & Avg. & 95.5 & 95.5 \\
\midrule
\multirow{6}{*}{Session 4} & \multirow{3}{*}{ProCA-B}
    & PCD Acc. & 90 & 100 \\
    & & TCD Acc. & 81.8 & 83.3 \\
    & & Avg. & 85.9 & \textbf{91.7} \\
\cmidrule(lr){2-5}
    & \multirow{3}{*}{GROTO}
    & PCD Acc. & 100 & 90 \\
    & & TCD Acc. & 90.9 & 75 \\
    & & Avg. & \textbf{95.5} & 82.5 \\
\midrule
\multirow{6}{*}{Session 5} & \multirow{3}{*}{ProCA-B}
    & PCD Acc. & 100 & 100 \\
    & & TCD Acc. & 90.9 & 90.9 \\
    & & Avg. & 95.5 & \textbf{95.5} \\
\cmidrule(lr){2-5}
    & \multirow{3}{*}{GROTO}
    & PCD Acc. & 100 & 100 \\
    & & TCD Acc. & 100 & 76.9 \\
    & & Avg. & \textbf{100} & 88.5 \\
\midrule
\multirow{6}{*}{Session 6} & \multirow{3}{*}{ProCA-B}
    & PCD Acc. & 100 & 90 \\
    & & TCD Acc. & 100 & 69.2 \\
    & & Avg. & 100 & \textbf{79.6} \\
\cmidrule(lr){2-5}
    & \multirow{3}{*}{GROTO}
    & PCD Acc. & 100 & 90 \\
    & & TCD Acc. & 100 & 64.3 \\
    & & Avg. & \textbf{100} & 77.2 \\
\midrule
\multirow{6}{*}{Session 7} & \multirow{3}{*}{ProCA-B}
    & PCD Acc. & 100 & 100 \\
    & & TCD Acc. & 100 & 83.3 \\
    & & Avg. & 100 & \textbf{91.7} \\
\cmidrule(lr){2-5}
    & \multirow{3}{*}{GROTO}
    & PCD Acc. & 100 & 100 \\
    & & TCD Acc. & 100 & 66.7 \\
    & & Avg. & \textbf{100} & 83.4 \\
\midrule
\multirow{6}{*}{Session 8} & \multirow{3}{*}{ProCA-B}
    & PCD Acc. & 100 & 90 \\
    & & TCD Acc. & 100 & 81.8 \\
    & & Avg. & \textbf{100} & \textbf{85.9} \\
\cmidrule(lr){2-5}
    & \multirow{3}{*}{GROTO}
    & PCD Acc. & 100 & 100 \\
    & & TCD Acc. & 90.9 & 71.4 \\
    & & Avg. & 95.5 & 85.7 \\
\bottomrule
\end{tabular}
\label{Table 8}
\end{table}

\begin{table*}[!htbp]
\centering
\footnotesize
\setlength{\tabcolsep}{5.7pt}
\caption{Mined accuracy (\%) comparisons of the HKPCM module with the ProCA-B method on Office-Home-CI.}
\begin{tabular}{llccccccccccccc}
\toprule
Session & Method & Metric & A$\rightarrow$C & A$\rightarrow$P & A$\rightarrow$R & C$\rightarrow$A & C$\rightarrow$P & C$\rightarrow$R & P$\rightarrow$A & P$\rightarrow$C & P$\rightarrow$R & R$\rightarrow$A & R$\rightarrow$C & R$\rightarrow$P \\
\midrule
\multirow{6}{*}{Session 1} & \multirow{3}{*}{ProCA-B} & PCD Acc. & 90 & 90 & 100 & 90 & 100 & 100 & 80 & 80 & 100 & 80 & 100 & 100 \\
& & TCD Acc. & 69.2 & 81.8 & 83.3 & 90 & 66.7 & 83.3 & 100 & 72.7 & 90.9 & 66.7 & 83.3 & 83.3 \\
& & Avg. & 79.6 & 85.9 & 91.7 & 90 & 83.4 & 91.7 & 90 & 76.4 & 95.5 & 73.4 & \textbf{91.7} & \textbf{91.7} \\
\cmidrule(lr){2-15}
& \multirow{3}{*}{GROTO} & PCD Acc. & 100 & 100 & 100 & 100 & 100 & 100 & 100 & 90 & 100 & 100 & 100 & 100 \\
& & TCD Acc. & 62.5 & 90.9 & 90.9 & 100 & 71.4 & 83.3 & 100 & 75 & 90.9 & 90.9 & 66.7 & 76.9 \\
& & Avg. & \textbf{81.3} & \textbf{95.5} & \textbf{95.5} & \textbf{100} & \textbf{85.7} & \textbf{91.7} & \textbf{100} & \textbf{82.5} & \textbf{95.5} & \textbf{95.5} & 83.4 & 88.5 \\
\midrule
\multirow{6}{*}{Session 2} & \multirow{3}{*}{ProCA-B} & PCD Acc. & 90 & 100 & 100 & 90 & 100 & 100 & 100 & 90 & 100 & 90 & 100 & 100 \\
& & TCD Acc. & 75 & 100 & 100 & 100 & 90.9 & 100 & 83.3 & 47.4 & 100 & 81.8 & 100 & 100 \\
& & Avg. & \textbf{82.5} & \textbf{100} & \textbf{100} & \textbf{95} & 95.5 & 100 & 91.7 & 68.7 & 100 & 85.9 & \textbf{100} & \textbf{100} \\
\cmidrule(lr){2-15}
& \multirow{3}{*}{GROTO} & PCD Acc. & 90 & 100 & 90 & 90 & 100 & 100 & 100 & 90 & 100 & 100 & 100 & 100 \\
& & TCD Acc. & 52.9 & 90.9 & 90 & 90 & 90.9 & 100 & 90.9 & 50 & 100 & 90.9 & 55.6 & 83.3 \\
& & Avg. & 71.5 & 95.5 & 90 & 90 & \textbf{95.5} & \textbf{100} & \textbf{95.5} & \textbf{70} & \textbf{100} & \textbf{95.5} & 77.8 & 91.7 \\
\midrule
\multirow{6}{*}{Session 3} & \multirow{3}{*}{ProCA-B} & PCD Acc. & 90 & 90 & 100 & 90 & 100 & 100 & 80 & 90 & 100 & 100 & 100 & 100 \\
& & TCD Acc. & 45 & 69.2 & 76.9 & 75 & 90.9 & 90.9 & 80 & 47.4 & 90.9 & 50 & 100 & 100 \\
& & Avg. & \textbf{67.5} & 79.6 & 88.5 & 82.5 & \textbf{95.5} & \textbf{90.9} & 80 & \textbf{68.7} & \textbf{95.5} & 75 & \textbf{100} & \textbf{100} \\
\cmidrule(lr){2-15}
& \multirow{3}{*}{GROTO} & PCD Acc. & 80 & 90 & 100 & 90 & 100 & 100 & 80 & 80 & 100 & 100 & 80 & 100 \\
& & TCD Acc. & 50 & 69.2 & 76.9 & 81.8 & 83.3 & 71.4 & 80 & 50 & 66.7 & 76.9 & 47.1 & 90.9 \\
& & Avg. & 65 & \textbf{79.6} & \textbf{88.5} & \textbf{85.9} & 91.7 & 85.7 & 80 & 65 & 83.4 & \textbf{88.5} & 63.6 & 95.5 \\
\midrule
\multirow{6}{*}{Session 4} & \multirow{3}{*}{ProCA-B} & PCD Acc. & 100 & 100 & 100 & 100 & 100 & 100 & 100 & 100 & 100 & 100 & 100 & 100 \\
& & TCD Acc. & 52.6 & 90.9 & 100 & 90.9 & 100 & 100 & 90.9 & 45.5 & 100 & 66.7 & 90.9 & 90.9 \\
& & Avg. & 76.3 & \textbf{95.5} & 100 & 95.5 & \textbf{100} & 100 & 95.5 & 72.8 & 100 & 83.4 & \textbf{95.5} & \textbf{95.5} \\
\cmidrule(lr){2-15}
& \multirow{3}{*}{GROTO} & PCD Acc. & 100 & 100 & 100 & 100 & 100 & 100 & 100 & 100 & 100 & 100 & 100 & 100 \\
& & TCD Acc. & 62.5 & 83.3 & 100 & 100 & 83.3 & 100 & 90.9 & 66.7 & 100 & 90.9 & 62.5 & 83.3 \\
& & Avg. & \textbf{81.3} & 91.7 & \textbf{100} & \textbf{100} & 91.7 & \textbf{100} & \textbf{95.5} & \textbf{83.4} & \textbf{100} & \textbf{95.5} & 81.3 & 91.7 \\
\midrule
\multirow{6}{*}{Session 5} & \multirow{3}{*}{ProCA-B} & PCD Acc. & 90 & 100 & 100 & 90 & 100 & 100 & 90 & 90 & 100 & 90 & 100 & 100 \\
& & TCD Acc. & 60 & 100 & 100 & 90 & 100 & 100 & 81.8 & 75 & 100 & 81.8 & 100 & 100 \\
& & Avg. & 75 & \textbf{100} & 100 & \textbf{90} & \textbf{100} & 100 & \textbf{85.9} & \textbf{82.5} & 100 & \textbf{85.9} & \textbf{100} & 100 \\
\cmidrule(lr){2-15}
& \multirow{3}{*}{GROTO} & PCD Acc. & 100 & 100 & 100 & 90 & 100 & 100 & 90 & 90 & 100 & 100 & 90 & 100 \\
& & TCD Acc. & 62.5 & 90.9 & 100 & 81.8 & 83.3 & 100 & 75 & 50 & 100 & 71.4 & 69.2 & 100 \\
& & Avg. & \textbf{81.3} & 95.5 & \textbf{100} & 85.9 & 91.7 & \textbf{100} & 82.5 & 70 & \textbf{100} & 85.7 & 79.6 & \textbf{100} \\
\midrule
\multirow{6}{*}{Session 6} & \multirow{3}{*}{ProCA-B} & PCD Acc. & 100 & 90 & 100 & 90 & 100 & 90 & 100 & 100 & 100 & 80 & 100 & 100 \\
& & TCD Acc. & 62.5 & 81.8 & 90.9 & 81.8 & 83.3 & 81.8 & 83.3 & 52.6 & 100 & 44.4 & 83.3 & 83.3 \\
& & Avg. & \textbf{81.3} & 85.9 & \textbf{95.5} & 85.9 & \textbf{91.7} & 81.8 & \textbf{91.7} & 76.3 & \textbf{100} & 62.2 & \textbf{91.7} & 91.7 \\
\cmidrule(lr){2-15}
& \multirow{3}{*}{GROTO} & PCD Acc. & 100 & 90 & 100 & 100 & 90 & 90 & 100 & 100 & 100 & 100 & 80 & 100 \\
& & TCD Acc. & 50 & 81.8 & 83.3 & 90.9 & 75 & 81.8 & 76.9 & 55.6 & 76.9 & 90.9 & 44.4 & 83.3 \\
& & Avg. & 75 & \textbf{85.9} & 91.7 & \textbf{95.5} & 82.5 & \textbf{85.9} & 88.5 & \textbf{77.8} & 88.5 & \textbf{95.5} & 62.2 & \textbf{91.7} \\
\bottomrule
\end{tabular}
\label{Table 9}
\end{table*}

In this section, we conduct more comparisons with the ProCA-B to examine the effectiveness of the hybrid knowledge-driven positive class mining~(HKPCM) module through two evaluation metrics~(PCD Acc. and TCD Acc.) as shown in ~\Cref{Table 8}, ~\Cref{Table 9} and ~\Cref{Table 10}. The PCD Acc. is the mined truly positive classes divided by the number of ground-truth positive classes, and the TCD Acc. is the mined truly positive classes divided by the number of all mined positive classes. The higher PCD Acc. means that the module mines more positive classes, and the higher TCD Acc. means that the module mines fewer false positive classes. With both metrics, we can evaluate the model's ability to discover positive classes. 

\captionsetup[table]{aboveskip = 3.5pt}
\begin{table}[!htbp]
\centering
\footnotesize  
\setlength{\tabcolsep}{1pt}  
\caption{Mined accuracy~(\%) comparisons of the HKPCM module with the ProCA-B method on Office-31-CI.}
\begin{tabular}{llccccccc}
\toprule
Session & Method & Metric & A$\rightarrow$D & A$\rightarrow$W & D$\rightarrow$A & D$\rightarrow$W & W$\rightarrow$A & W$\rightarrow$D \\
\midrule
\multirow{6}{*}{Session 1} & \multirow{3}{*}{ProCA-B} & PCD Acc. & 90.0 & 90.0 & 90.0 & 100.0 & 100.0 & 100.0 \\
& & TCD Acc. & 100.0 & 100.0 & 75.0 & 100.0 & 76.9 & 100.0 \\
& & Avg. & 95.0 & 95.0 & 82.5 & 100.0 & 88.5 & 100.0 \\
\cmidrule(lr){2-9}
& \multirow{3}{*}{GROTO} & PCD Acc. & 100.0 &100.0 & 90.0 & 100.0 & 90.0 & 100.0 \\
& & TCD Acc. & 100.0 & 100.0 & 90.0 & 100.0 & 90.0 & 100.0 \\
& & Avg. & \textbf{100.0} & \textbf{100.0} & \textbf{90.0} & \textbf{100.0} & \textbf{90.0} & \textbf{100.0} \\
\midrule
\multirow{6}{*}{Session 2} & \multirow{3}{*}{ProCA-B} & PCD Acc. & 100.0 & 100.0 & 90.0 & 100.0 & 100.0 & 100.0 \\
& & TCD Acc. & 83.3 & 76.9 & 69.2 & 100.0 & 76.9 & 100.0 \\
& & Avg. & 91.7 & 88.5 & 79.6 & 100.0 & 88.5 & 100.0 \\
\cmidrule(lr){2-9}
& \multirow{3}{*}{GROTO} & PCD Acc. & 100.0 & 100.0 & 90.0 & 100.0 & 100.0 & 100.0 \\
& & TCD Acc. & 83.3 & 100.0 & 75.0 & 100.0 & 83.3 & 100.0 \\
& & Avg. & \textbf{91.7} & \textbf{100.0} & \textbf{82.5} & \textbf{100.0} & \textbf{91.7} & \textbf{100.0} \\
\midrule
\multirow{6}{*}{Session 3} & \multirow{3}{*}{ProCA-B} & PCD Acc. & 100.0 & 100.0 & 100.0 & 100.0 & 100.0 & 100.0 \\
& & TCD Acc. & 100.0 & 100.0 & 83.3 & 100.0 & 83.3 & 100.0 \\
& & Avg. & 100.0 & 100.0 & 91.7 & 100.0 & 91.7 & 100.0 \\
\cmidrule(lr){2-9}
& \multirow{3}{*}{GROTO} & PCD Acc. & 100.0 & 100.0 & 100.0 & 100.0 & 100.0 & 100.0 \\
& & TCD Acc. & 90.9 & 100.0 & 76.9 & 100.0 & 76.9 & 100.0 \\
& & Avg. & \textbf{95.5} & \textbf{100.0} & \textbf{88.5} & \textbf{100.0} & \textbf{88.5} & \textbf{100.0} \\
\bottomrule
\end{tabular}
\label{Table 10}
\end{table}

\section{More analysis of PTD module} 
\label{Section 5}

\captionsetup[table]{aboveskip = 4pt}
\begin{table}[!htbp]
\centering
\footnotesize  
\setlength{\tabcolsep}{3.5pt} 
\caption{Average classification accuracy~(\%) comparisons of 0 to 9 Classes across sessions on Office-31-CI.}
\begin{tabular}{cccccccccccc}
\hline
Task & Method & U & CI & SF & Session 1 & Session 2 & Session 3 & Avg. \\
\hline
\multirow{4}{*}{A$\rightarrow$D} & 
LCFD & \cmark & \xmark & \cmark & 87.3 & 44.2 & 30.0 & 53.8 \\
& CIDA* & \xmark & \cmark & \cmark & 90.3 & 77.1 & 70.4 & 79.3 \\
& DIFO & \cmark & \xmark & \cmark & 98.9 & 98.5 & 99.3 & 98.9 \\
& GROTO & \cmark & \cmark & \cmark & \textbf{100.0} & \textbf{100.0} & \textbf{100.0} & \textbf{100.0} \\
\hline
\multirow{4}{*}{A$\rightarrow$W} &
LCFD & \cmark & \xmark & \cmark & 54.5 & 52.5 & 56.8 & 54.6 \\
& CIDA* & \xmark & \cmark & \cmark & 82.1 & 70.6 & 64.5 & 72.4 \\
& DIFO & \cmark & \xmark & \cmark & 98.4 & 95.8 & 98.5 & 97.6 \\
& GROTO & \cmark & \cmark & \cmark & \textbf{100.0} & \textbf{100.0} & \textbf{100.0} & \textbf{100.0} \\
\hline
\multirow{4}{*}{D$\rightarrow$A} & 
LCFD & \cmark & \xmark & \cmark & 73.7 & 46.9 & 42.3 & 54.3 \\
& CIDA* & \xmark & \cmark & \cmark & 71.0 & 63.9 & 48.1 & 61.0 \\
& DIFO & \cmark & \xmark & \cmark & 74.2 & 62.6 & 66.4 & 67.7 \\
& GROTO& \cmark & \cmark & \cmark & \textbf{84.8} & \textbf{80.3} & \textbf{80.7} & \textbf{81.9} \\
\hline
\multirow{4}{*}{D$\rightarrow$W} & 
LCFD & \cmark & \xmark & \cmark & 71.9 & 48.4 & 49.0 & 56.4 \\
& CIDA* & \xmark & \cmark & \cmark & 97.4 & 99.8 & 95.1 & 97.4 \\
& DIFO & \cmark & \xmark & \cmark & 75.2 & 78.3 & 77.0 & 76.8 \\
& GROTO & \cmark & \cmark & \cmark & \textbf{100.0} & \textbf{100.0} & \textbf{100.0} & \textbf{100.0} \\
\hline
\multirow{4}{*}{W$\rightarrow$A} & 
LCFD & \cmark & \xmark & \cmark & 70.8 & 48.1 & 40.1 & 53.0 \\
& CIDA* & \xmark & \cmark & \cmark & 72.1 & 65.7 & 52.7 & 63.5 \\
& DIFO & \cmark & \xmark & \cmark & 79.6 & 63.8 & 72.7 & 72.0 \\
& GROTO & \cmark & \cmark & \cmark & \textbf{84.8} & \textbf{81.3} & \textbf{80.1} & \textbf{82.1} \\
\hline
\multirow{4}{*}{W$\rightarrow$D} & 
LCFD & \cmark & \xmark & \cmark & 60.6 & 52.3 & 71.9 & 61.6 \\
& CIDA* & \xmark & \cmark & \cmark & 100.0 & 97.7 & 98.8 & 98.8 \\
& DIFO & \cmark & \xmark & \cmark & 77.4 & 80.2 & 80.5 & 79.3 \\
& GROTO & \cmark & \cmark & \cmark & \textbf{100.0} & \textbf{100.0} & \textbf{100.0} & \textbf{100.0} \\
\hline
\end{tabular}
\label{Table 12}
\end{table}

\begin{figure*}[!htbp]
\centering
\includegraphics[width=2.1\columnwidth]{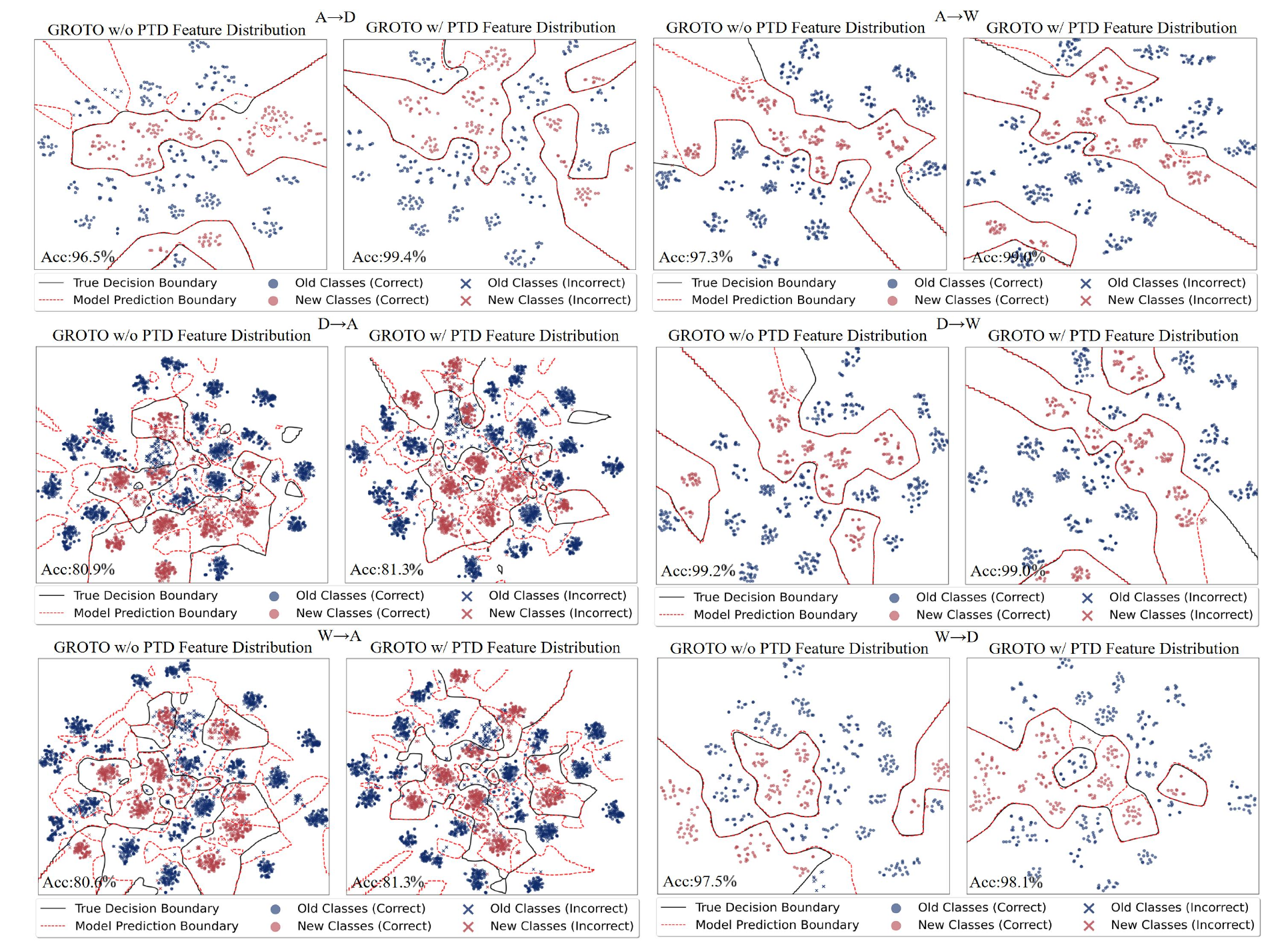} 
\caption{Feature distribution and decision boundary visualizations of ``GROTO w/o PTD" and ``GROTO w/ PTD" models on Office-31-CI.}
\label{Figure 7}
\end{figure*}

\begin{figure*}[!htbp]
\centering
\includegraphics[width=2.1\columnwidth]{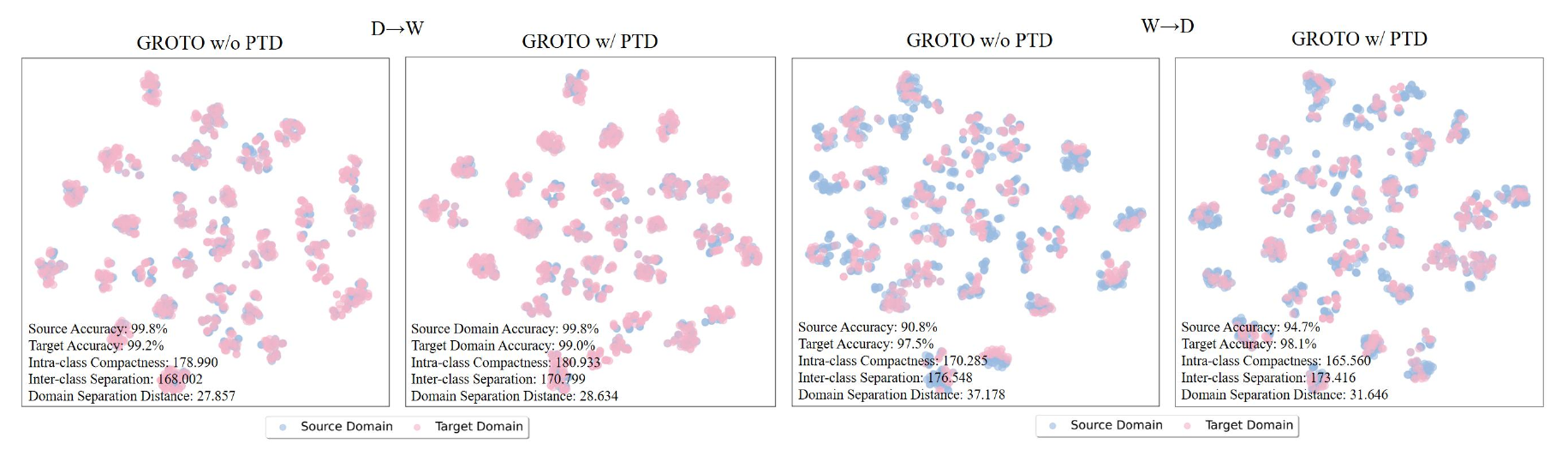} 
\caption{Source and target feature distribution visualizations of ``GROTO w/o PTD" and ``GROTO w/ PTD" models.}
\label{Figure 8}
\end{figure*}

In this section, we first compare GROTO with the class-incremental domain adaptation method~(\textit{i.e.}, CIDA~\cite{kundu2020class}), and source-free universal domain adaptation method~(\textit{e.g.}, DIFO~\cite{tang2024sourcefreedomainadaptationfrozen} and LCFD~\cite{tang2024unified}) to evaluate the average classification performance of 0 to 9 Classes across sessions on Office-31-CI and Office-Home-CI 
in ~\Cref{Table 12} and ~\Cref{Table 13}. Then, we show more feature distribution and decision boundary visualizations of the trained 30 classes about ``GROTO w/o PTD" and ``GROTO w/ PTD" models on Office-31-CI in ~\Cref{Figure 7}. To further analyze the effectiveness of the PTD module, we also visualize the source and target feature distribution of the trained 30 classes when respectively targeting the D and W domain on Office-31-CI, and compute the five metrics~(\textit{i.e.}, the source accuracy, target accuracy, intra-class compactness, inter-class separation, and domain separation distance) of ``GROTO w/o PTD" and ``GROTO w/ PTD" models in ~\Cref{Figure 8}.

The 0 to 9 classes are the training classes at the first incremental session, the model focuses excessively on the new classes due to the shortage of old-class data when learning new classes at later sessions. Therefore, the optimizer updates weights to adjust towards the features and patterns of new classes, resulting in shocks and disturbances to the knowledge of old classes~(\textit{e.g.}, the class indexes of 0 to 9). It can be seen that the 0 to 9 old class classification performance of the GROTO model at different incremental sessions are almost entirely better than the compared CIDA, DIFO, and LCFD methods in ~\Cref{Table 12} and ~\Cref{Table 13}. Moreover, the method in the class-incremental scenario~(\textit{i.e.}, CIDA) outperforms some of the source-free universal domain adaptation methods for the 0 to 9 old class classification accuracy. It also illustrates the importance of continually mitigating the disturbances of new target knowledge to old ones in the CI-SFUDA problem.

\begin{table*}[htbp]
\centering
\footnotesize  
\setlength{\tabcolsep}{9.5pt}  
\caption{Average classification accuracy~(\%) comparisons of 0 to 9 Classes across sessions on Office-Home-CI.}
\begin{tabular}{cccccccccccc}
\hline
Task & Method & U & CI & SF & Session 1 & Session 2 & Session 3 & Session 4 & Session 5 & Session 6 & Avg. \\
\hline
\multirow{4}{*}{A$\rightarrow$C} & CIDA* & \xmark & \cmark & \cmark & 50.5 & 45.1 & 40.4 & 37.3 & 34.8 & 32.2 & 40.1 \\
& DIFO & \cmark & \xmark & \cmark & 63.0 & 29.9 & 26.7 & 28.8 & 13.8 & 29.5 & 32.0 \\
& LCFD & \cmark & \xmark & \cmark & 56.3 & 43.4 & 48.2 & 44.1 & 46.3 & 45.5 & 47.3 \\
& GROTO & \cmark & \cmark & \cmark & \textbf{74.6} & \textbf{72.6} & \textbf{70.0} & \textbf{66.8} & \textbf{68.6} & \textbf{66.1} & \textbf{69.8} \\
\hline
\multirow{4}{*}{A$\rightarrow$P} & CIDA* & \xmark & \cmark & \cmark & 66.2 & 59.6 & 57.0 & 52.0 & 50.8 & 45.9 & 55.2 \\
& DIFO & \cmark & \xmark & \cmark & 81.5 & 45.3 & 47.9 & 50.7 & 42.7 & 52.2 & 52.9 \\
& LCFD & \cmark & \xmark & \cmark & 75.3 & 59.8 & 65.3 & 68.1 & 65.5 & 64.0 & 66.3 \\
& GROTO & \cmark & \cmark & \cmark & \textbf{88.9} & \textbf{87.7} & \textbf{85.5} & \textbf{84.9} & \textbf{84.8} & \textbf{85.0} & \textbf{86.1} \\
\hline
\multirow{4}{*}{A$\rightarrow$R} & CIDA* & \xmark & \cmark & \cmark & 67.6 & 60.5 & 60.2 & 58.7 & 58.3 & 49.1 & 59.1 \\
& DIFO & \cmark & \xmark & \cmark & 82.6 & 67.7 & 73.3 & 57.7 & 60.9 & 58.5 & 66.8 \\
& LCFD & \cmark & \xmark & \cmark & 76.6 & 73.0 & 75.7 & 77.9 & 79.1 & 76.0 & 76.4 \\
& GROTO& \cmark & \cmark & \cmark & \textbf{90.4} & \textbf{88.0} & \textbf{87.7} & \textbf{86.2} & \textbf{86.0} & \textbf{84.9} & \textbf{87.2} \\
\hline
\multirow{4}{*}{C$\rightarrow$A} & CIDA* & \xmark & \cmark & \cmark & 41.8 & 41.6 & 37.1 & 35.2 & 35.4 & 36.5 & 37.9 \\
& DIFO & \cmark & \xmark & \cmark & 77.9 & 57.0 & 53.7 & 63.2 & 54.2 & 52.7 & 59.8 \\
& LCFD & \cmark & \xmark & \cmark & 72.3 & 62.5 & 66.9 & 65.5 & 63.5 & 62.6 & 65.6 \\
& GROTO & \cmark & \cmark & \cmark & \textbf{85.9} & \textbf{83.8} & \textbf{83.1} & \textbf{82.2} & \textbf{81.2} & \textbf{82.0} & \textbf{83.0} \\
\hline
\multirow{4}{*}{C$\rightarrow$P} & CIDA* & \xmark & \cmark & \cmark & 60.1 & 54.2 & 53.9 & 50.0 & 50.0 & 48.6 & 52.8 \\
& DIFO & \cmark & \xmark & \cmark & \textbf{81.9} & 30.3 & 47.6 & 44.8 & 35.1 & 49.6 & 48.2 \\
& LCFD & \cmark & \xmark & \cmark & 54.6 & 37.4 & 37.7 & 39.5 & 35.8 & 42.0 & 41.2 \\
& GROTO & \cmark & \cmark & \cmark & 81.2 & \textbf{79.8} & \textbf{78.4} & \textbf{79.7} & \textbf{78.4} & \textbf{78.2} & \textbf{79.3} \\
\hline
\multirow{4}{*}{C$\rightarrow$R} & CIDA* & \xmark & \cmark & \cmark & 64.9 & 50.4 & 52.7 & 52.1 & 50.8 & 46.6 & 52.9 \\
& DIFO & \cmark & \xmark & \cmark & 80.0 & 59.8 & 66.9 & 51.4 & 55.0 & 58.4 & 61.9 \\
& LCFD & \cmark & \xmark & \cmark & 74.9 & 75.4 & 75.2 & 71.6 & 73.9 & 71.5 & 73.8 \\
& GROTO & \cmark & \cmark & \cmark & \textbf{90.8} & \textbf{88.9} & \textbf{88.4} & \textbf{85.0} & \textbf{85.4} & \textbf{85.3} & \textbf{87.3} \\
\hline
\multirow{4}{*}{P$\rightarrow$A} & CIDA* & \xmark & \cmark & \cmark & 49.1 & 48.9 & 48.5 & 50.0 & 50.2 & 51.6 & 49.7 \\
& DIFO & \cmark & \xmark & \cmark & 79.1 & 56.0 & 49.1 & 53.9 & 53.2 & 53.1 & 57.4 \\
& LCFD & \cmark & \xmark & \cmark & 76.9 & 59.3 & 61.8 & 57.6 & 57.3 & 63.0 & 62.6 \\
& GROTO & \cmark & \cmark & \cmark & \textbf{87.4} & \textbf{82.2} & \textbf{85.5} & \textbf{84.8} & \textbf{83.4} & \textbf{84.0} & \textbf{84.5} \\
\hline
\multirow{4}{*}{P$\rightarrow$C} & CIDA* & \xmark & \cmark & \cmark & 50.7 & 42.8 & 39.4 & 36.2 & 36.6 & 33.5 & 39.9 \\
& DIFO & \cmark & \xmark & \cmark & 58.1 & 25.5 & 23.7 & 31.8 & 20.2 & 23.6 & 30.5 \\
& LCFD & \cmark & \xmark & \cmark & 54.6 & 37.4 & 37.7 & 39.5 & 35.8 & 42.0 & 41.2 \\
& GROTO & \cmark & \cmark & \cmark & \textbf{72.0} & \textbf{68.3} & \textbf{64.1} & \textbf{65.2} & \textbf{64.9} & \textbf{63.7} & \textbf{66.3} \\
\hline
\multirow{4}{*}{P$\rightarrow$R} & CIDA* & \xmark & \cmark & \cmark & 65.0 & 62.3 & 61.4 & 59.4 & 59.7 & 59.0 & 61.1 \\
& DIFO & \cmark & \xmark & \cmark & 81.0 & 68.4 & 71.4 & 57.0 & 59.3 & 65.1 & 67.0 \\
& LCFD & \cmark & \xmark & \cmark & 76.8 & 76.7 & 78.4 & 77.7 & 76.8 & 75.8 & 77.0 \\
& GROTO & \cmark & \cmark & \cmark & \textbf{91.4} & \textbf{88.9} & \textbf{88.4} & \textbf{86.5} & \textbf{87.0} & \textbf{85.3} & \textbf{87.9} \\
\hline
\multirow{4}{*}{R$\rightarrow$A}& CIDA* & \xmark & \cmark & \cmark & 62.3 & 63.5 & 61.9 & 62.9 & 62.4 & 64.0 & 62.8 \\
& DIFO & \cmark & \xmark & \cmark & 81.9 & 55.3 & 61.3 & 63.4 & 59.6 & 58.4 & 63.3 \\
& LCFD & \cmark & \xmark & \cmark & 72.9 & 67.6 & 68.7 & 69.0 & 70.1 & 68.0 & 69.4 \\
& GROTO & \cmark & \cmark & \cmark & \textbf{90.6} & \textbf{86.3} & \textbf{86.1} & \textbf{86.2} & \textbf{84.9} & \textbf{85.3} & \textbf{86.6} \\
\hline
\multirow{4}{*}{R$\rightarrow$C} & CIDA* & \xmark & \cmark & \cmark & 44.6 & 45.4 & 41.3 & 41.2 & 40.0 & 38.0 & 41.7 \\
& DIFO & \cmark & \xmark & \cmark & 61.9 & 23.1 & 23.6 & 28.6 & 17.3 & 22.8 & 29.5 \\
& LCFD & \cmark & \xmark & \cmark & 55.7 & 41.2 & 40.7 & 41.1 & 36.8 & 40.8 & 42.7 \\
& GROTO & \cmark & \cmark & \cmark & \textbf{73.8} & \textbf{68.8} & \textbf{69.6} & \textbf{69.6} & \textbf{68.4} & \textbf{67.1} & \textbf{69.5} \\
\hline
\multirow{4}{*}{R$\rightarrow$P} & CIDA* & \xmark & \cmark & \cmark & 71.6 & 69.6 & 68.3 & 66.4 & 66.2 & 65.1 & 67.9 \\
& DIFO & \cmark & \xmark & \cmark & 82.4 & 42.9 & 52.5 & 52.8 & 43.0 & 36.6 & 51.7 \\
& LCFD & \cmark & \xmark & \cmark & 73.0 & 68.1 & 69.9 & 69.8 & 58.9 & 69.8 & 68.3 \\
& GROTO & \cmark & \cmark & \cmark & \textbf{89.2} & \textbf{88.0} & \textbf{88.1} & \textbf{86.7} & \textbf{87.1} & \textbf{86.9} & \textbf{87.7} \\
\hline
\end{tabular}
\label{Table 13}
\end{table*}

The PTD module allows the classifier weights of the target positive classes at each session to point-to-point distill towards the corresponding weights of the source model that well-trained in all classes. 
The ~\Cref{Figure 7} shows that the PTD module can mitigate the over-adaptation of the model weights to the new classes, alleviating the model tendency of misclassifying the old class data into the new classes. And then it makes the model decision boundary of the old and new classes smoother and closer to the true ones, thus improving the classification accuracy.

In the further visualizations in ~\Cref{Figure 8}, we can see that due to the domain gap, although the PTD module mitigates the impact of the new-class knowledge on the old ones, the target model needs to maintain feature representations that are compatible with the source classifier weights in the target feature space, while maintaining the separability of the different class data on the target domain. 
This trade-off may induce the extractor to extract target features with a certain degree of focusing more on those feature patterns similar to the source domain instead of the most discriminative features of the target domain itself. Therefore, despite the fluctuation of intra-class compactness in some subtasks, the spacing between different classes is effectively broadened, and the forgetting of source knowledge is mitigated. It also illustrates that this slight sacrifice is beneficial, because it enables the model to focus more on cross-domain knowledge migration and incremental knowledge integration, rather than limiting itself to target intra-class compactness and inter-class separability. 
When facing the target domain data, although the target model does not completely fit the optimal feature patterns of the target domain, it provides clearer boundaries and spaces for the integration of new classes, avoiding classifying the old-class data to the new classes. Simultaneously, it can alleviate the shocks of the target knowledge to the source ones to a certain extent, thus achieving more stable and efficient performance.

\section{More session accuracy comparisons}
\label{Section 6}
In this section, we present the session accuracy of CIDA~\cite{kundu2020class}, ProCA-B~\cite{lin2022prototype}, LCFD~\cite{tang2024unified}, DIFO~\cite{tang2024sourcefreedomainadaptationfrozen}, and LEAD-B~\cite{qu2024lead} on ImageNet-Caltech-CI in ~\Cref{Table 14}. It illustrates that GROTO outperforms other methods, which proves that GROTO can perform well for both old and new target classes at each session.

\begin{table*}[!htbp]
\centering
\caption{Session accuracy~(\%) comparisons in the class-incremental scenario on ImageNet-Caltech-CI. U, CI, and SF respectively represent unsupervised, class-incremental, and source data-free. -B indicates the backbone is changed to ViT-B. * indicates the result is derived from ProCA.
}
{\fontsize{9}{10}\selectfont
\resizebox{\textwidth}{!}{
\fontsize{9}{10}\selectfont
\begin{tabular}{l@{\hskip 20pt}c@{\hskip 10pt}c@{\hskip 10pt}c@{\hskip 10pt}|ccccccccc}
\toprule
\multirow{2}{*}{Method} & \multirow{2}{*}{U} & \multirow{2}{*}{CI} & \multirow{2}{*}{SF} & \multicolumn{9}{c}{ImageNet-Caltech-CI} \\
\cmidrule(lr){5-13}
& & & & Session 1 & Session 2 & Session 3 & Session 4 & Session 5 & Session 6 & Session 7 & Session 8 & Avg. \\
\midrule
CIDA*~(ECCV20) & \xmark & \cmark & \cmark & 58.6 & 61.3 & 65.4 & 67.1 & 65.9 & 69.4 & 68.8 & 69.3 & 65.7 \\
ProCA-B~(ECCV22) & \cmark  & \cmark & \xmark & 85.9 & 83.4 & 88.3 & 87.2 & 88.3 & 87.4 & 87.8 & 87.8 & 87.0 \\
LCFD~(Arxiv24) & \cmark & \xmark & \cmark & 44.4 & 57.3 & 62.8 & 65.9 & 67.7 & 71.2 & 72.8 & 72.1 & 64.3 \\
DIFO~(CVPR24) & \cmark & \xmark & \cmark & 67.3 & 33.5 & 48.0 & 51.8 & 62.1 & 52.6 & 58.9 & 61.4 & 54.4 \\
LEAD-B~(CVPR24) & \cmark & \xmark & \cmark & 85.3 & 84.2 & 88.4 & 88.4 & 89.7 & 87.1 & 82.7 & 71.1 & 84.6 \\
\midrule
GROTO~(Ours) & \cmark & \cmark & \cmark & \textbf{91.8} & \textbf{88.8} & \textbf{91.7} & \textbf{90.8} & \textbf{91.5} & \textbf{89.4} & \textbf{89.2} & \textbf{88.8} & \textbf{90.3} \\
\bottomrule
\label{Table 14}
\end{tabular}
}}
\end{table*}

\section{More details on the training of GROTO}
\label{Section 7}
In this section, we provide more details of the GROTO training. We perform the algorithm on the RTX 3090 GPU and use the SGD optimizer with the weight decay, momentum, and learning rate respectively set to $1\times10^{-6}$, 0.9, and 0.001. When the number of iterations at each session is less than 5, we use the source model to identify multi-granularity class prototypes and balance the number of prototypes for different positive classes. We first count the number of prototypes for each positive class to obtain the minimum value $p$. For each positive class, we select $p$ prototypes based on the prediction confidence in descending order. 
In the memory bank, we update the representative exemplars every 10 iterations for Office-31-CI and ImageNet-Caltech-CI, and every 15 iterations for Office-Home-CI. 
To simulate virtual distributional shifts and enhance the consistent learning capability across multiple datasets, we adopt the general weak augmentations~(\textit{e.g.} random crop, random grayscale, random horizontal flip, and color jitter) instead of the dataset-specific or manually designed ones~\cite{shorten2019survey}.

\section{More visualization comparisons}
\label{Section 8}
\begin{figure*}[!htbp]
\centering
\includegraphics[width=2\columnwidth]{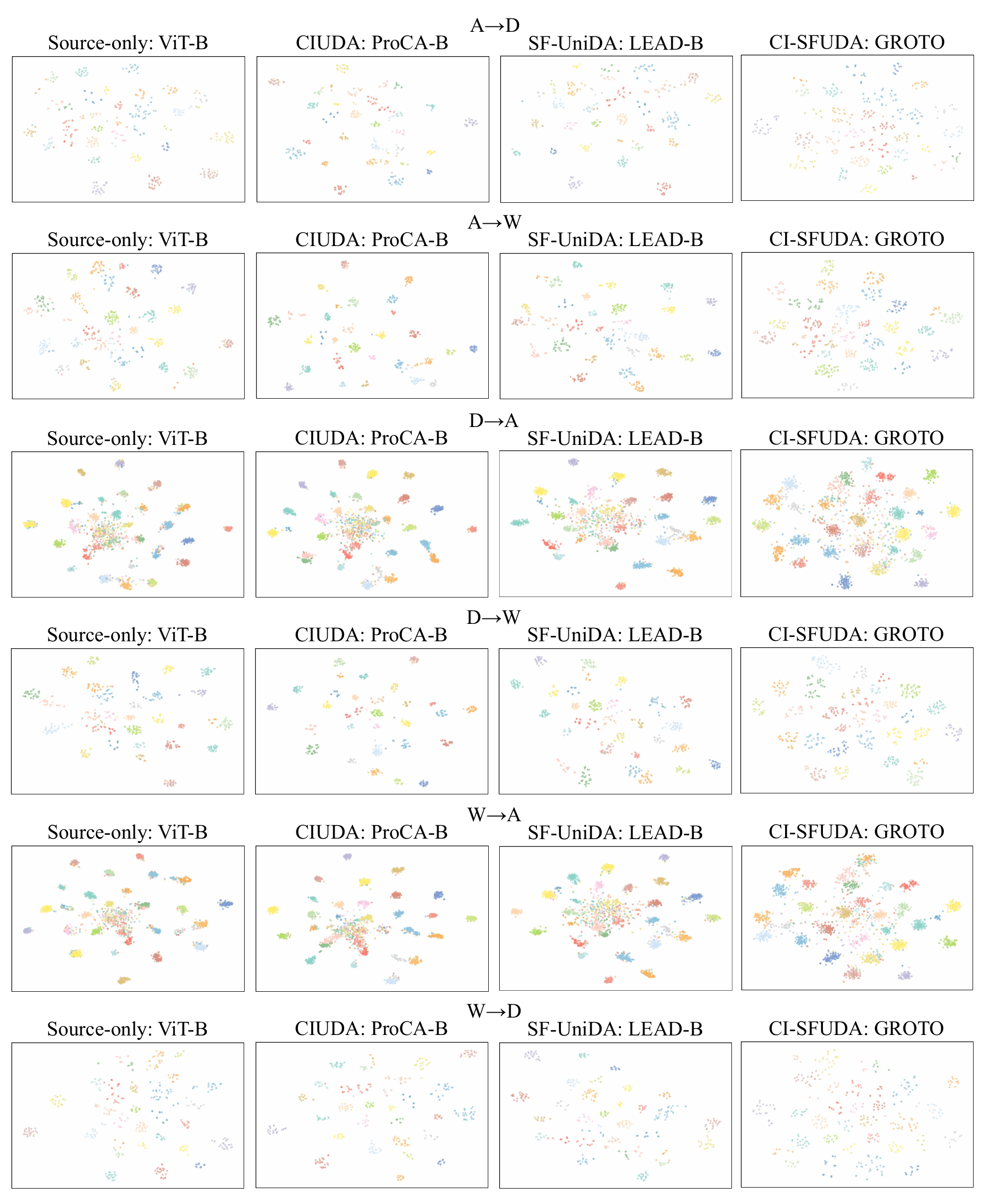} 
\caption{Target feature distribution of source-only~(ViT-B), ProCA-B, LEAD-B, and GROTO methods on Office-31-CI.}
\label{Figure 9}
\end{figure*}

In this section, we provide more t-SNE visualization comparisons for ViT-B~\cite{dosovitskiy2020image}), ProCA-B~\cite{lin2022prototype}, LEAD-B~\cite{qu2024lead} and GROTO methods as shown in ~\Cref{Figure 9}.

\section{The constructions of incremental datasets}
\label{Section 9}
In this section, we show the constructions of the three incremental datasets in ~\Cref{Table 15} and ~\Cref{Table 16}. We follow the same dataset setting as ProCA~\cite{lin2022prototype}, allowing GROTO can be more fairly compared to other methods.

\begin{table}[!htbp]
\centering
\footnotesize  
\caption{Class names of the Office-31-CI and Office-Home-CI.}
\begin{tabular}{@{}ccm{4cm}@{}}
\toprule
Dataset & Session & \multicolumn{1}{c}{Class Names} \\
\midrule
\multirow{3}{*}[-1cm]{Office-31-CI} & Session 1 & Back Pack, Bike, Bike Helmet, Bookcase, Bottle, Calculator, Desk Chair, Desk Lamp, Desktop Computer, File Cabinet \\
& Session 2 & Headphones, Keyboard, Laptop Computer, Letter Tray, Mobile Phone, Monitor, Mouse, Mug, Paper Notebook, Pen \\
& Session 3 & Phone, Printer, Projector, Punchers, Ring Binder, Ruler, Scissors, Speaker, Stapler, Tape Dispenser \\
\midrule
\multirow{6}{*}[-1.7cm]{Office-Home-CI} & Session 1 & Drill, Exit Sign, Bottle, Glasses, Computer, File Cabinet, Shelf, Toys, Sink, Laptop \\
& Session 2 & Kettle, Folder, Keyboard, Flipflops, Pencil, Bed, Hammer, ToothBrush, Couch, Bike \\
& Session 3 & Postit Notes, Mug, Webcam, Desk Lamp, Telephone, Helmet, Mouse, Pen, Monitor, Mop \\
& Session 4 & Sneakers, Notebook, Backpack, Alarm Clock, Push Pin, Paper Clip, Batteries, Radio, Fan, Ruler \\
& Session 5 & Pan, Screwdriver, Trash Can, Printer, Speaker, Eraser, Bucket, Chair, Calendar, Calculator \\
& Session 6 & Flowers, Lamp Shade, Spoon, Candles, Clipboards, Scissors, TV, Curtains, Fork, Soda \\
\bottomrule
\end{tabular}
\label{Table 15}
\vspace{-4pt}
\end{table}

\begin{table}[!htbp]
\centering
\footnotesize  
\caption{Class indices at each session on ImageNet-Caltech-CI.}
\begin{tabular}{@{}cm{3.1cm}m{3.1cm}@{}}
\toprule
Session & Class indices of C$\rightarrow$I & Class indices of I$\rightarrow$C \\
\midrule
{Session 1} & [0, 2, 7, 9, 11, 27, 28, 29, 30, 33] & [1, 9, 24, 39, 51, 69, 71, 79, 94, 99] \\
\midrule
{Session 2} & [37, 39, 40, 44, 45, 47, 50, 60, 62, 68] & [112, 113, 145, 148, 171, 288, 308, 311, 314, 315] \\
\midrule
{Session 3} & [71, 75, 76, 82, 85, 86, 87, 88, 89, 90] & [327, 334, 340, 354, 355, 361, 366, 367, 413, 414] \\
\midrule
{Session 4} & [92, 94, 96, 97, 106, 107, 108, 109, 110, 112] & [417, 433, 441, 447, 471, 472, 479, 506, 508, 514] \\
\midrule
{Session 5} & [114, 115, 116, 123, 126, 128, 133, 134, 141, 145] & [543, 546, 555, 560, 566, 571, 574, 579, 593, 594] \\
\midrule
{Session 6} & [146, 150, 151, 157, 160, 163, 165, 170, 172, 177] & [604, 605, 620, 621, 637, 651, 664, 671, 713, 745] \\
\midrule
{Session 7} & [178, 179, 181, 185, 188, 192, 193, 196, 198, 200] & [760, 764, 779, 784, 805, 806, 814, 839, 845, 849] \\
\midrule
{Session 8} & [209, 211, 215, 219, 225, 227, 228, 229, 230, 234] & [852, 859, 870, 872, 876, 879, 895, 907, 910, 920] \\
\bottomrule
\end{tabular}
\label{Table 16}
\end{table}

\section{More complexity analysis} 
\label{section3}
In this section, we first analyze the computational complexity of GROTO, including both the ViT-B backbone network complexity and the complexity of our proposed modules. Simultaneously, we provide a complexity analysis for each module. Then we compare the training costs with the CIUDA method~(\textit{i.e.}, ProCA-B~\cite{lin2022prototype}).

\subsection{Overall complexity analysis of GROTO}

We first analyze the computation complexity of ViT-B/16 we used in GROTO. For each transformer block, the multi-head self-attention requires $O(4ND^2 + 2N^2D)$ computations, where the query, key and value projections take $O(3ND^2)$, attention computation needs $O(2N^2D)$, and output projection costs $O(ND^2)$. The MLP block with two linear layers~(the expansion ratio of 4) requires $O(8ND^2)$ computations. Therefore, a single Transformer block has a total complexity of $O(12ND^2 + 2N^2D)$. For the complete ViT-B/16 with $L=12$ layers processing 224×224 images~($N=196$ and $D=768$), the total backbone theoretical complexity is $O(L(12ND^2 + 2N^2D))$.

We then analyze the computation complexity of each module in ~\Cref{Table 6}, the increased training time and computation complexity are mainly derived from the HKPCM and PTFS modules.

\begin{table}[!htbp]
\centering  
\footnotesize
\setlength{\tabcolsep}{3.5pt}
\caption{The complexity analysis on Office-31-CI~(A$\rightarrow$D). The complexity is for a single image, and time is for a module.}
\begin{tabular}{cccccc}  
\toprule            
\multicolumn{1}{c}{Module} & \multicolumn{1}{c}{HKPCM} & \multicolumn{1}{c}{PTFS} & \multicolumn{1}{c}{PTD} & \multicolumn{1}{c}{KR} \\ 
\midrule
Complexity & $O(n_tK^2)$ & $O(n_t|P|+B+B^2)$ & $O(4N^2)$ & $O(n_tM)$ \\
Time(s) & 3.44 & 2.81 & 7e-4 & 0.20 \\
\bottomrule
\end{tabular}
\label{Table 6}
\end{table}

\subsection{Training Cost Comparison}

As shown in ~\Cref{Table 7}, the parameter number of GROTO is identical to ProCA-B, because these modules do not introduce extra parameters. The slightly increased GPU memory~(0.5G) is mainly used to store the augmented data and intermediate features. The increased training time~(0.28h) is primarily due to the computation of HKPCM and PTFS modules.

\begin{table}[!htbp]
\centering
\footnotesize
\setlength{\tabcolsep}{5pt}
\centering
\caption{The training cost on Office-31-CI~(A$\rightarrow$D).}
\begin{tabular}{c|cccc}
\toprule
Metric & Flops(G) & \#Param(M) & Memory(G) & Training time(h) \\
\midrule
ProCA-B & 16.86 & 85.67 & 17.94 & 0.23 \\
GROTO & 16.86 & 85.67 & 18.44 & 0.51 \\
\bottomrule
\end{tabular}
\label{Table 7}
\end{table}

{
    \small
    \bibliographystyle{ieeenat_fullname}
    \bibliography{main}
}

\end{document}